\pdfoutput=1

\documentclass[11pt]{article}

\usepackage{acl}

\usepackage{times}
\usepackage{latexsym}

\usepackage[T1]{fontenc}

\usepackage[utf8]{inputenc}

\usepackage{microtype}

\usepackage{amsmath,amssymb,amsfonts}
\usepackage{algorithmic}
\usepackage{graphicx}
\usepackage{multirow}
\usepackage{textcomp}
\usepackage{subcaption}
\usepackage{booktabs}
\usepackage{longtable}
\usepackage{supertabular}
\usepackage{geometry}

\def\L\mathcal{L}

\def\v{\mathbf{v}}

\def\L{\mathcal{L}}

\def\I{\mathbf{I}}

\def\I{\mathbf{I}}

\def\o{\mathbf{o}} 
\def\q{\mathbf{q}} 

\def\0{\mathbf{0}}
\def\I{\mathbf{I}}

\def\L{\mathcal{L}}

\title{Interpretable Medical Image Visual Question Answering via Multi-Modal Relationship Graph Learning}

\author{\normalfont \textbf{Xinyue Hu}\textsuperscript{1}, \textbf{Lin Gu}\textsuperscript{2,3}, \textbf{Kazuma Kobayashi}\textsuperscript{4}, \textbf{Qiyuan An}\textsuperscript{1}, \textbf{Qingyu Chen}\textsuperscript{5}, \\ \textbf{Zhiyong Lu}\textsuperscript{5}, \textbf{Chang Su}\textsuperscript{6},\textbf{ Tatsuya Harada}\textsuperscript{2,3}, \textbf{Yingying Zhu}\textsuperscript{1} \\
  \textsuperscript{1}The University of Texas Arlington, USA, \textsuperscript{2}RIKEN, Japan\\
  \textsuperscript{3}University of Tokyo, Japan \\
  \textsuperscript{4}National Cancer Center Research Institute, Japan\\
  \textsuperscript{5}National Library of Medicine - National Institutes of Health, USA \\
  \textsuperscript{6}Temple University, USA \\
  \\}

\begin{document}
\maketitle
\begin{abstract}
Medical visual question answering (VQA) aims to answer clinically relevant questions regarding input medical images. This technique has the potential to improve the efficiency of medical professionals while relieving the burden on the public health system, particularly in resource-poor countries. Existing medical VQA methods tend to encode medical images and learn the correspondence between visual features and questions without exploiting the spatial, semantic, or medical knowledge behind them. This is partially because of the small size of the current medical VQA dataset, which often includes simple questions. Therefore, we first collected a comprehensive and large-scale medical VQA dataset, focusing on chest X-ray images. The questions involved detailed relationships, such as disease names, locations, levels, and types in our dataset. Based on this dataset, we also propose a  novel baseline method by constructing three different relationship graphs: spatial relationship, semantic relationship, and implicit relationship graphs on the image regions, questions, and semantic labels. The answer and graph reasoning paths are learned for different questions. 
\end{abstract}

\section{Introduction}
\begin{figure*}[h]
    \centering
    \includegraphics[width=0.76\textwidth]{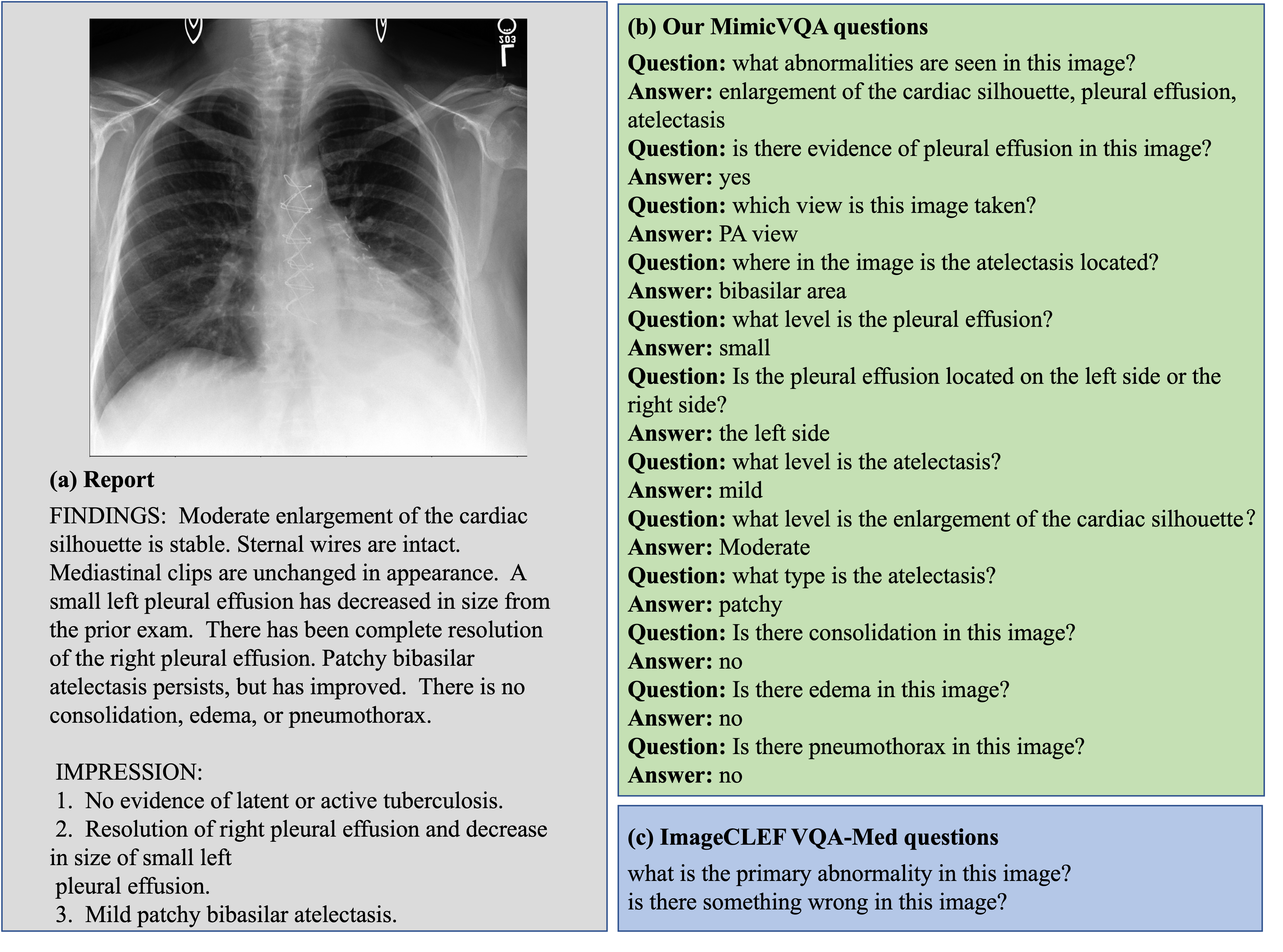}
    \caption{A comparison between our constructed VQA dataset and the existing ImageCLEF VQA-Med dataset. (a) The report corresponds to the given Chest X-ray image. (b) Our constructed question settings, including \textit{abnormality}, \textit{presence}, \textit{view}, \textit{location}, \textit{level}, and \textit{type}. (c) The design of the ImageCLEF VQA-MED questions is too simple.}
    \label{fig:fig1}
\end{figure*}
\begin{figure*}
    \centering
    \includegraphics[width=0.76\textwidth]{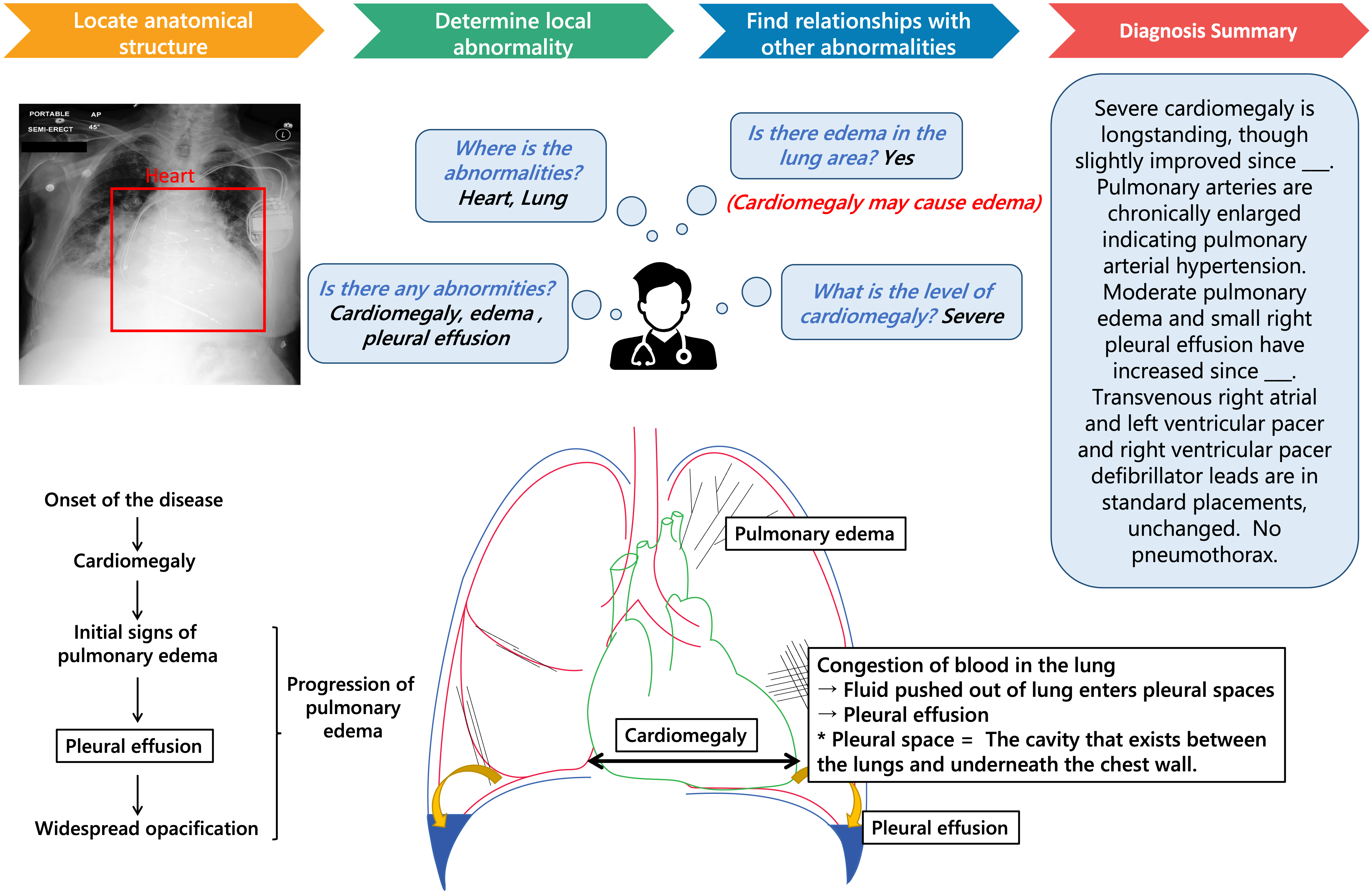}
    \caption{Clinical motivation for the construction of our dataset and VQA method}
    \label{fig:motivattion}
\end{figure*}

Medical visual question answering (VQA) is a technique that answers clinically relevant questions regarding a medical image. This is a challenging task that requires both medical image diagnosis and natural language understanding. Medical VQA can provide clinicians with a "second opinion" in interpreting medical images and decrease the risk of misdiagnosis~\cite{Tschandl2020HumancomputerCF}. It also has the potential to relieve the burden on radiologists by partially taking over their expert consultant role to answer questions from physicians and patients, preventing the disruption of their workflow and improving efficiency \cite{LinSurveyMVQA22}. 

Artificial intelligence (AI) can be utilized to perform these tasks, which can assist in reducing global health inequalities in low- and middle-income countries. For example, when interpreting complex cases, the second opinion provided by the medical VQA system may significantly enhance the junior clinicians' confidence when specialized experts are not available.  Deploying such a system would also alleviate the shortage of healthcare services in resource-poor regions, \textit{i.e.}, Africa, which is home to only 3\% of the world’s healthcare labor force and bears 25\% of the global disease burden~\cite{crisp2011global}. Medical VQA can contribute to sustainable development goals (SDGs) by reducing the cost of healthcare in resource-poor countries and promoting healthy living and well-being.

Most of the current medical VQA methods adopt a joint embedding framework~\cite{Antol_VQA} that relies on pre-trained convolutional neural networks (CNNs) as backbones, such as the VGGNet \cite{simonyan2014very}, to capture visual structures. These black-box models tend to exploit the dataset bias by capturing the superficial correlations among visual appearances, questions, and answers\cite{balanced_vqa_v2,Cao_2021_ICCV}. In fact, some state-of-the-art medical VQA algorithms do not even utilize the question feature and generate the answer using only the image feature \cite{LinSurveyMVQA22}. The disadvantage of over-reliance on training data only is particularly obvious in the medical domain because of the limited and diverse training data. 
A Multiple Meta-model Quantifying(MMQ) process to utilize meta-learning to improve performance on small-sized datasets was proposed in ~\cite{mmq}. However, for larger datasets, the improvement is limited.


More critically, current medical VQA datasets have several limitations: 1) They mostly focus on very simple questions such as "What is the abnormality in this image?" or "Is there something wrong in the image?" (Fig.~\ref{fig:fig1} (c))~\cite{ImageCLEF-VQA-Med2021}. 2) They cover a wide range of modalities (MRI, CT, and X-ray) and various body sites (neuroimaging, chest X-rays, and abdominal CT/MRI scans).
As the pathology of diseases in different body parts is very complicated and heterogeneous, medical images along with questions differ markedly across modalities, specialties, and diseases. Therefore, a universal VQA model is not a panacea and cannot be generalized to different modalities and body locations.

%

In the progression of a disease, multiple diseases may be interconnected. For instance, as shown in Fig.~\ref{fig:motivattion}. cardiomegaly (enlargement of the heart) can increase pressure on the lungs, leading to initial signs of pulmonary edema (fluid in the lungs). This fluid can then accumulate in the pleural spaces, causing pleural effusion (fluid in the pleural spaces). 
Therefore, during the diagnostic process, doctors typically follow a "coarse-to-fine" routine. 
They first locate the relevant anatomical structure (such as the heart), then determine local abnormality (such as cardiomegaly), find relationships with other abnormalities (such as pleural effusion and edema), and finally make a diagnosis summary.
Based on this, we constructed a dataset focusing on chest X-ray images with comprehensive questions on abnormalities, body location, disease level, abnormality type, and evidence to mimic the process of practical diagnosis. Fig.~\ref{fig:fig1} (b) shows examples of question-answer pairs in the dataset.

To build this dataset, we first extracted a KeyInfo dataset, which contains the key information of a report, such as abnormalities, attributes, and the relationships between them. Then, we constructed the question-answer pairs based on the information collected from the KeyInfo dataset.  

\begin{figure*}[h]
    \centering
    \includegraphics[width=.85\textwidth]{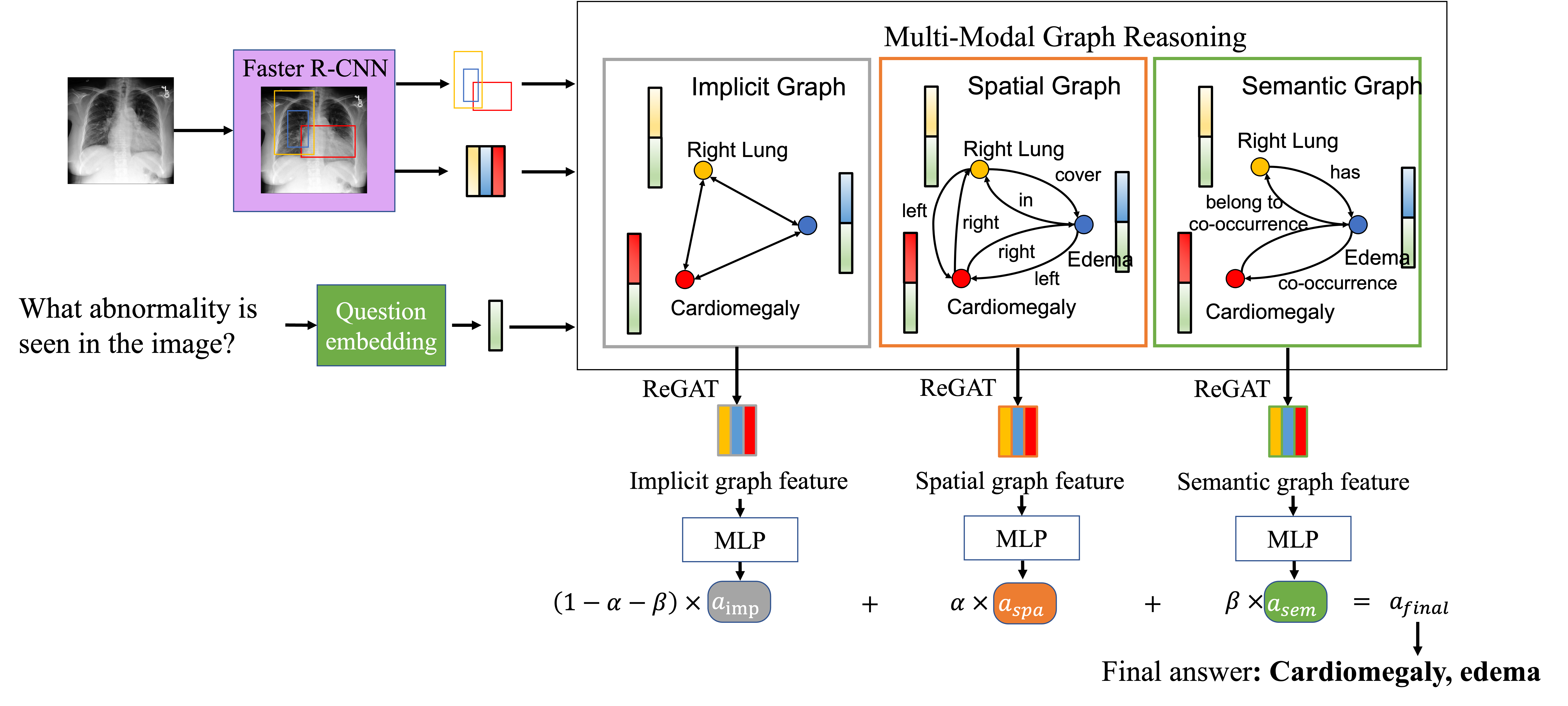}
    \caption{Proposed Multi-Modal Graph Learning Medical VQA Framework.}
    \label{fig:overview}
\end{figure*}

In addition, to mimic the process of "Find relationships with other abnormalities" and enhance the generality of practical situations, we proposed a novel medical VQA framework that can understand and deeply combine expert knowledge and diagnostic reasoning to provide interpretable and reliable AI systems to be used in real clinical settings.
This is the first framework that explicitly leverages rigorous medical knowledge graphs and considers the spatial relations between anatomical structures and diseased regions, as shown in Fig.~\ref{fig:overview}. 
Our contributions can be summarized as:
1) We constructed a specific, comprehensive, and challenging medical VQA dataset focusing on chest X-ray image analysis with detailed questions on diseases, body parts, levels, and types. 

2) We proposed a novel multi-relationship graph model, which leverages visual, spatial, and semantic relationships for the VQA task. The semantic relationship is built based on the knowledge graph of anatomical structures and diseases. 

3) The learned graph model can also interpret the reasoning path of how the visual question is answered. \textbf{The code and dataset will be released upon publication.} 

\section{Related Work}

Previous visual question answering (VQA) methods trained the convolutional neural network (CNN) and long short-term memory (LSTM) based architectures in an end-to-end manner~\cite{Xu2016AskAA,Shin16CVPR}. Subsequently, the joint embedding structure has become prevalent~\cite{Antol_VQA,yu2017mfb}, which is widely adopted as a baseline method~\cite{LinSurveyMVQA22}. Stemming from the general-domain VQA, the medical VQA~\cite{LinSurveyMVQA22} has undergone rapid development owing to the emergence of various medical datasets \cite{liu2021slake,ImageCLEF-VQA-Med2021,he2020pathvqa,lau2018dataset}. Among them, most of the methods also employ joint embedding to capture the relationship between the image and question. However, it has been argued that the existing methods tend to leverage superficial correlations rather than a deep understanding of the image~\cite{balanced_vqa_v2,Cao_2021_ICCV}.  Some methods \cite{zhou2018employing,anderson2018bottom,jiang2018pythia} simply feed medical question-answer(QA) datasets into existing VQA models, without considering the relationships between anatomical structures and findings in radiology images. For example, in \cite{zhan2020medical}, the focus was on distinguishing question types; however, the learning of high-level features from radiology images was not emphasized. Prevailing visual and textual models pre-trained on general datasets were exploited to extract both features in \cite{abacha2018nlm,zhou2018employing}. 
A general-domain VQA explores an "adult-level common sense" to support  inference~\cite{WU2017Survey}. However, reading the medical images and answering the clinical-specific questions requires professional knowledge and experience. To fill this gap,  we introduce a novel multi-modal graph-learning method to leverage expert knowledge, and spatial and semantic relationships for medical VQA.

\section{Method}


\textbf{Our Method Overview.}
Given an input medical image $\I_i$ and a question $\q_i$, as shown in Fig. ~\ref{fig:overview}, we aim to predict the answer to $\q_i$ based on image information. We propose a multimodal graph-learning model, as shown in Fig.~\ref{fig:overview}, by first extracting the region of interest (ROI) using a pre-trained Faster R-CNN and considering each ROI as a node in the graph.
We considered three different relationships to build the graph relationship/edges: 1) spatial relationships based on ROI-wise spatial locations, 2) semantic relationships based on medical expert knowledge, and 3) implicit relationships to discover additional latent relationships. We then compute the answer by fusing multimodal graphs with a multilayer perceptron network.

\textbf{Detection of Anatomy and Disease Location}
As shown in Fig.~\ref{fig:KG} in the Appendix, we propose to introduce the knowledge of anatomical structures and diseases by first locating their positions, or ROIs.
We employed a Faster R-CNN~\cite{ren2015faster} on the labeled dataset to train the detection model for anatomical structures and diseases, using the MIMIC chest X-ray~\cite{mimic} and VinDr~\cite{nguyen2020vindr} datasets, respectively. After locating these regions, we extracted the visual features using a Faster R-CNN \cite{ren2015faster} for each ROI.
The detected ROIs and their image features are denoted by $\{\o_{i}\}_{i=1}^N$, where $\o_i \in \mathbb{R}^{d_o}$ is the visual feature of one detected ROI, $N$ is the total number of detected ROIs. 


\subsection{Multi-Modal Graph Construction.}
As shown in Fig.~\ref{fig:overview}, we constructed the following three modal graphs after extracting the anatomical and disease ROIs:  
1) Spatial relation graph 
2) Semantic relation graph and 
3) Implicit relation graph. 
The visual graph is defined as $G=\{\mathcal{V}, \mathcal{E}_{spa}, \mathcal{E}_{sem}, \mathcal{E}_{imp}\}$, 
Each vertex feature $\v_i \in \mathcal{V}$ is defined as 
$\v_i = [\o_i\|\q] \in \mathbb{R}^{d_o+d_q} \mathrm{for} \ i = 1, \dots, N $,
where $\mathcal{E}_{spa}$, $\mathcal{E}_{sem}$ and $\mathcal{E}_{imp}$ are the sets of the spatial, semantic, and implicit edges, $N$ is the number of vertices, $\|$ represents concatenation,
$\q \in \mathbb{R}^{d_q}$ is the embedded question. 
To embed questions $\q$, we followed the procedure of ~\cite{regat,norcliffe2018learning} to tokenize and embed each word with GloVe~\cite{pennington2014glove} before feeding them into a bidirectional GRU~\cite{cho2014learning}. 

\textbf{Spatial Graph.}
In the spatial relation graph, we define the spatial relationship following a previous study ~\cite{regat} to include 11 types of spatial relations between detected ROIs (such as inside (class1) and cover (class2))~\cite{yao2018exploring}. 
The edge label between the node $i$ and the node $j$ is defined as $c_{lab(i,j)} = r$, where $r$ is the class of the relationship, $r = 1, 2, \cdots, K$, K is the number of spatial relationship classes, which is 11.
When $d_{ij} > t$, we set $c_{lab(i,j)}=0$, where $d_{ij}$ is the Euclidean distance between the center points of the bounding boxes corresponding to the nodes $i$ and node $j$, and $t$ is the threshold.

\textbf{Semantic Graph.}
In line with the desire to improve collaboration between AI experts and clinicians, we define two types of semantic relationships \cite{zhang2020radiology,zhou2021contrast,lian2021structure} in our semantic relationship graph:
    \textit{1) {Anatomical Knowledge graph}}. 
    Following a previous study~\cite{zhang2020radiology}, we constructed an anatomical knowledge graph to model the body parts and disease relationships, as shown in Fig.~\ref{fig:anaKG} in the Appendix. We refined the original knowledge graph to better suit our task by removing irrelevant nodes and adding more relevant ones. The newly added nodes are highlighted in red. The nodes in the solid and dashed boxes represent disease labels and anatomical structures, respectively.
 \textit{2) Co-occurrence Knowledge graph}. 
    Following \cite{zhou2021contrast,lian2021structure}, we defined a disease co-occurrence knowledge graph as shown in Fig.~\ref{fig:coKG} in the Appendix. 
    The co-occurrence relationship was extracted by counting and normalizing the co-occurrence frequency of different disease labels from the clinical note dataset~\cite{mimic}. We connect these two nodes in the graph when $c_{ij} > t$, where $t$ is a threshold, and $c_{ij}$ means the co-occurrence frequency between the i-th and the j-th node.

To apply the knowledge graphs in Fig.~\ref{fig:KG} into our model, we assigned all detected ROIs to a combined graph of both the anatomical knowledge graph and the co-occurrence knowledge graphs.
Each ROI corresponds to a node in the knowledge graphs and is connected to the ROIs that correspond to all neighboring nodes in both knowledge graphs. 
The edge label $c_{lab(i,j)}$ in the anatomical knowledge graph was set to 1 in the adjacency matrix, whereas that in the co-occurrence knowledge graph was set to 2.

\textbf{Implicit Graph.}
In addition to the spatial and semantic relationships, we utilize the implicit relationships that have been demonstrated to be effective in general domain VQA problems for discovering latent relationships \cite{regat}. We followed the design of~\cite{regat} and used a fully connected graph to learn the implicit relationships between graph vertices. 

\textbf{Graph reasoning.} Please refer to the Appendix for the details.

\begin{table}[h]
\small
\caption{Full list of examples for each question type.}
\centering
\resizebox{0.5\textwidth}{!}{
\begin{tabular}{ll}
\toprule
type                & example                                               \\
\midrule
\multirow{4}{*}{Abnormality} & what abnormalities are seen in the image?            \\
                             & what abnormalities are seen in the \textit{\textlangle location\textrangle}?               \\
                             & is there any evidence of any abnormalities? \\
                             & is this image normal?                                 \\ \hline
\multirow{3}{*}{Presence}    & any evidence of \textit{\textlangle abnormality\textrangle}?               \\
                             & is there \textit{\textlangle abnormality\textrangle}?                                         \\
                             & is there \textit{\textlangle abnormality\textrangle} in the \textit{\textlangle location\textrangle}?                             \\ \hline
\multirow{3}{*}{View}        & which view is this image taken?                       \\
                             & is this PA view?                                      \\
                             & is this AP view?                                      \\ \hline
\multirow{4}{*}{Location}    & where  is the \textit{\textlangle abnormality\textrangle} located?                \\
                             & where is the \textit{\textlangle abnormality\textrangle}?                                     \\
                             & is the \textit{\textlangle abnormality\textrangle} located on the left/right?    \\
                             & is the \textit{\textlangle abnormality\textrangle} in the \textit{\textlangle location\textrangle}?                               \\ \hline
Level                        & what level is the \textit{\textlangle abnormality}?                                \\ \hline
Type                         & what type is the \textit{\textlangle abnormality\textrangle}?                                 \\ 
\bottomrule
\end{tabular}
}
\label{tab:question}
\end{table}

\section{Experiments}

\textbf{Experiments Setting.}
We train our model on our constructed dataset for 20 epochs with a batch size of 64 and with an Adam optimizer. 
The initial learning rate is 0.0005. We follow the setting of ~\cite{regat} by utilizing the warm-up strategy~\cite{goyal2017accurate}. 
The learning rate first increases to 0.002 at epoch 4, and then slowly decreases at epoch 15. 
The batch size is set to 64. 
We set 2 layers of relation-aware graph attention network for each graph.  The input feature dimension, hidden feature dimension, and output feature dimension are all set to 1024. The number of attention heads is set as 16. Each word is tokenized into a 600-dimension embedding(including 300-dimension GloVe embedding). The question embedding is obtained by feeding the embedded sequence word tokens into a one-layer GRU.
The experiments are conducted on PyTorch code using a GeForce RTX 3090 GPU. 
It takes 2 hours and 2 minutes to compute each graph for 20 epochs.
The demonstrated answers are chosen to be the top 4 answer predictions that have a higher score than 0.04.
We compare our method with one of the SOTAs conducted on the RAD-VQA dataset, MMQ~\cite{mmq},  which utilizes meta-learning to overcome the limited size of the training data.
The train/val/test sets are split sequentially in the ratio of 8:1:1.

\textbf{Existing Datasets.} 
Datasets for medical VQA are much smaller compared to general-domain VQA datasets, \textit{e.g.}, VQA v2 \cite{goyal2017accurate}, COCO-QA \cite{ren2015image}. For example, ImageCLEF VQA-Med-2019~\cite{abacha2019vqa} and VQA-RAD ~\cite{lau2018dataset} have only 4,200 images with 15,292 questions and 315 images with 3,515 questions, respectively, whereas general-domain datasets, such as COCO-QA, usually have more than 100,000 images and questions. Besides, the majority of questions in ImageCLEF VQA-Med and VQA-RAD are simple, close-ended, or multiple-choice questions, like "Is there something wrong in the image?" or "What is the primary abnormality in this image?".
SLAKE~\cite{liu2021slake} is a comprehensive dataset that introduces knowledge-based questions regarding CT, MRI, and X-ray modalities. Although concepts such as “the functionality of an organ,” “the cause of a disease,” or “the treatment of a disease” are involved, they are of limited types and with a limited number of questions. The dataset has only 642 images and 14,000 questions, where the questions are bilingual and include "vision-only" and "knowledge-based" types.

\begin{table}[h]
\centering
\caption{Comparison between the baseline model and our method with three relation graphs and the combined score. We used the AUC as the evaluation metric. AUC-micro computes the final AUC by aggregating the contributions of each class. AUC-macro treats all classes equally and computes the average AUC. "imp", "spa", and "sem" represent "implicit", "spatial", and "semantic", respectively.}
\label{tab:result}
\begin{tabular}{l|l|llll}
\multirow{2}{*}{AUC}          & \multirow{2}{*}{MMQ} & \multicolumn{4}{c}{Ours}                                            \\ \cline{3-6} 
                                &                      & imp & {spa} & \multicolumn{1}{l|}{sem} & all       \\ \hline
{micro}    & 0.981                & 0.995    & 0.995   & \multicolumn{1}{l|}{0.995}    & \textbf{0.996} \\ \hline
{macro} & 0.948                & 0.961   & 0.960   & \multicolumn{1}{l|}{0.957}    & \textbf{0.964}
\end{tabular}
\end{table}

The latest medical VQA dataset, ImageCLEF VQA-Med-2021~\cite{ImageCLEF-VQA-Med2021}, contains 5000 images and 5000 question-answer pairs, split into 4000, 500, and 500 for training, validation, and testing, respectively. There are five different imaging modalities: CT/MRI imaging, angiography, pathology, ultrasound, and diagnostic radiology. These images cover a large range of body structures, including the brain, chest, abdomen, arms, and legs. There were two types of questions: open-ended and closed-ended. Closed-ended questions ask whether the given image is normal or abnormal. The open-ended questions are diverse and include questions regarding the locations and types of abnormalities.
The key drawback of ImageCLEF is that it includes overcomplicated pathology images and a large number of diseases spanning a wide range of body parts. 

\textbf{Mimic-VQA Dataset.} 
To promote the development of VQA in the medical domain, we compiled a medical VQA dataset focusing on one modality and a specialty: the chest X-ray dataset. Nevertheless, our baseline model can be broadly generalized to different modalities and diseases. 
The VQA dataset includes three parts: image, question, and answer.
For the image set, we chose a large-scale MIMIC chest X-ray dataset~\cite{mimic} containing 227,835 studies and 377,110 images. 
Each study corresponds to one or more images, but only one report.
Fourteen finding labels were extracted for each study in~\cite{johnson2019mimic}. We further processed the MIMIC dataset by extracting fine-grained information from reports.

\begin{figure}[h]
    \centering
    \includegraphics[width=0.4\textwidth]{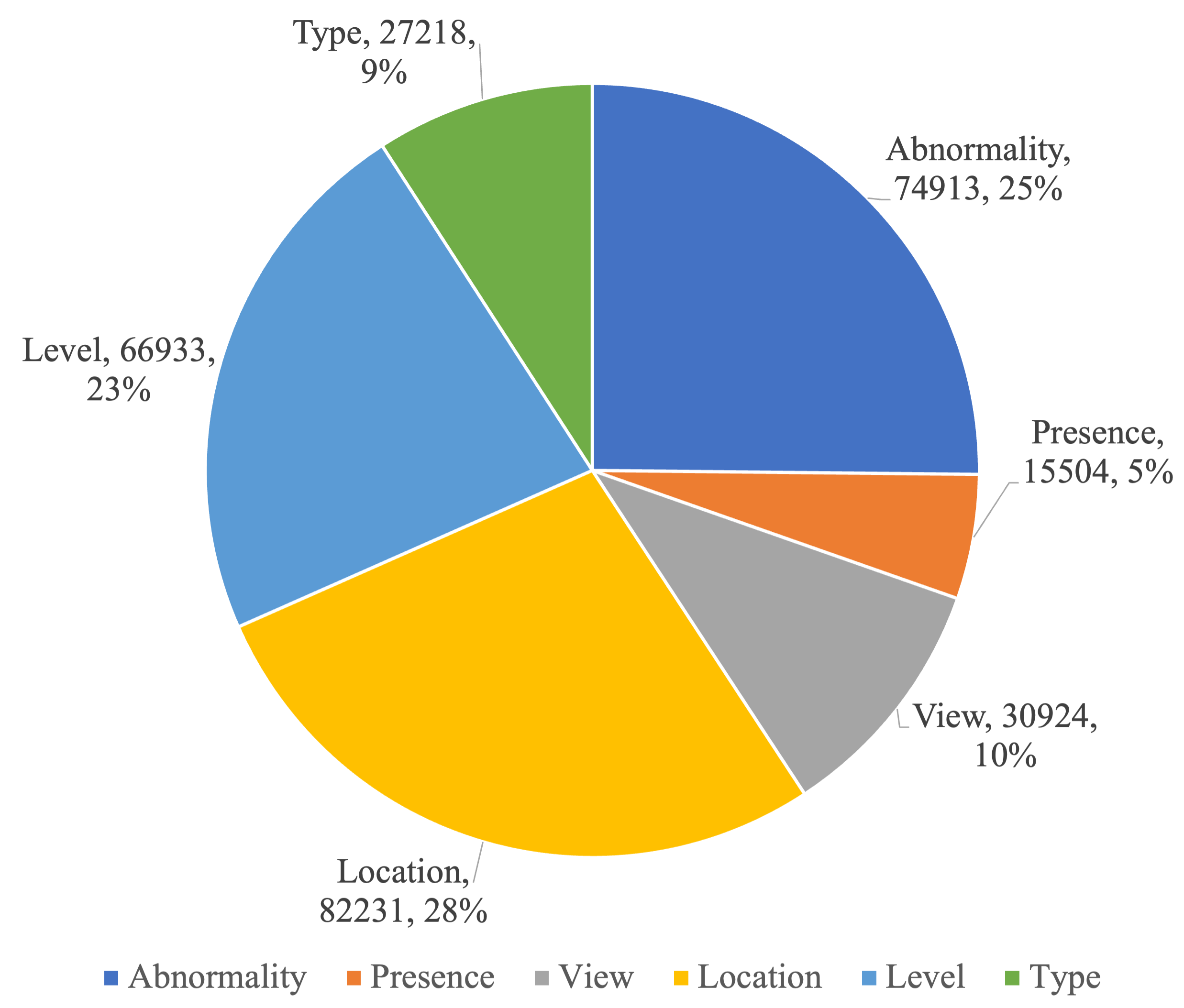}
    \caption{The statistics of each question type in the mimicVQA dataset.
    }
    \label{fig:stat}
\end{figure}

\textbf{Question Design.}
To cater to radiologists’ interests in disease diagnosis, our question design is an extension of the VQA-RAD question design.
It contains 11 types of questions on the following topics: \textit{abnormality}, \textit{presence}, \textit{modality}, \textit{organ}, \textit{other}, \textit{plane}, \textit{size}, \textit{count}, \textit{attribute}, \textit{color}, and \textit{position}. 
From these, we selected the four most relevant types of questions, \textit{abnormality}, \textit{presence}, \textit{view} ("plane" in VQA-RAD), and \textit{location} ("position" in VQA-RAD). In addition, we added \textit{type} and \textit{level} questions to our dataset. Table~\ref{tab:question} shows examples of each type of question in the dataset.
The VQA-RAD has only 315 images with 3,515 QA pairs.
Our mimic-VQA dataset significantly enlarged this number to 134,400 images and 297,723 QA pairs.
After filtering out some rare answers to alleviate the possible data imbalance problem, we obtained 169 answers.
The train/val/test sets are split in a ratio of 8:1:1.
Fig.~\ref{fig:stat} shows the statistics for each type of question in the dataset.

\textbf{Dataset Construction.}
To collect the information needed to generate QA pairs, we first constructed a KeyInfo dataset for the entire MIMIC dataset.
The KeyInfo dataset contains the key information of each report, such as abnormalities, and their corresponding locations, types, and levels.
We collected a list of abnormality keywords as well as lists of other attribute keywords, including location, level, and type.
\textbf{The list of all extractable abnormality keywords and the full list of attributes keywords can be found in the Appendix.}
Using regular expressions, we found the abnormality keywords in each report and searched for the corresponding attribute keywords that appeared before and after the abnormality keyword.
The regular expressions are defined by recursively validating the output labels and adjusting the regular expressions accordingly to minimize errors. The final validation results are shown in Section.~\ref{sec:validation}

After obtaining the keywords, we constructed a simple scene graph to establish their relationship and represent the report.
Please refer to the Appendix for a full list of abnormality and attribute keywords.
Thus, the KeyInfo dataset is constructed.
After constructing the KeyInfo dataset, we were able to obtain all the information needed to generate questions, including abnormalities, attributes, and the relationships between them.

\textbf{Dataset validation.}
\label{sec:validation}
To ensure the reliability of our constructed dataset, we had two human verifiers evaluate a randomly selected sample of 1700 question-answer pairs from the dataset. The results of this validation, shown in Table~\ref{tab:evaluation}, indicate that the overall accuracy of the Mimic-VQA dataset is 98.4\%, which is credible enough for training.

\begin{table}[h]
\caption{Validation results by human verifiers}
\label{tab:evaluation}
\begin{center}
\resizebox{\columnwidth}{!}{
\begin{tabular}{llll}
\hline
\multicolumn{1}{c}{\bf Verifier} & \multicolumn{1}{c}{\bf example \# } & \multicolumn{1}{c}{\bf correctness \# } & \multicolumn{1}{c}{\bf Acc} 
\\ \hline
Verifier 1                & 782                              & 772                                 & 98.72\%                   \\
Verifier 2                & 773                             & 762                                 & 98.57\%                  \\
Total                     & 1555                             & 1534                                 & 98.64\%                  \\\hline
\end{tabular}
}
\end{center}
\end{table}

\textbf{Quantitative results.}
Table~\ref{tab:result} presents a comparison between our model and the compared model. 
We also performed an ablation study to determine how different relationship graphs benefit from each other. 
It can be observed that our method outperforms the baseline model under both the AUC-micro and AUC-macro metrics.
Owing to the limited capability of meta-learning on large datasets, MMQ failed to demonstrate excessive performance on our mimic-VQA dataset.
In addition, the combined score is higher than the score of any single relation graph, proving that the combination of implicit, spatial, and semantic relation graphs compensates for each other's deficiencies.

\textbf{For details of the AUC score of each answer in our dataset, please refer to Table~\ref{tab:longtab} in the Appendix.}

Moreover, the results of our semantic graph, which was constructed based on the knowledge graphs, show an overall better performance than the other two graphs, suggesting that knowledge graphs are particularly helpful. In addition, the high combined scores indicate that the three different types of relationships can benefit from each other in answering questions.

\begin{figure*}[h]
    \centering
    \includegraphics[width=0.6\textwidth]{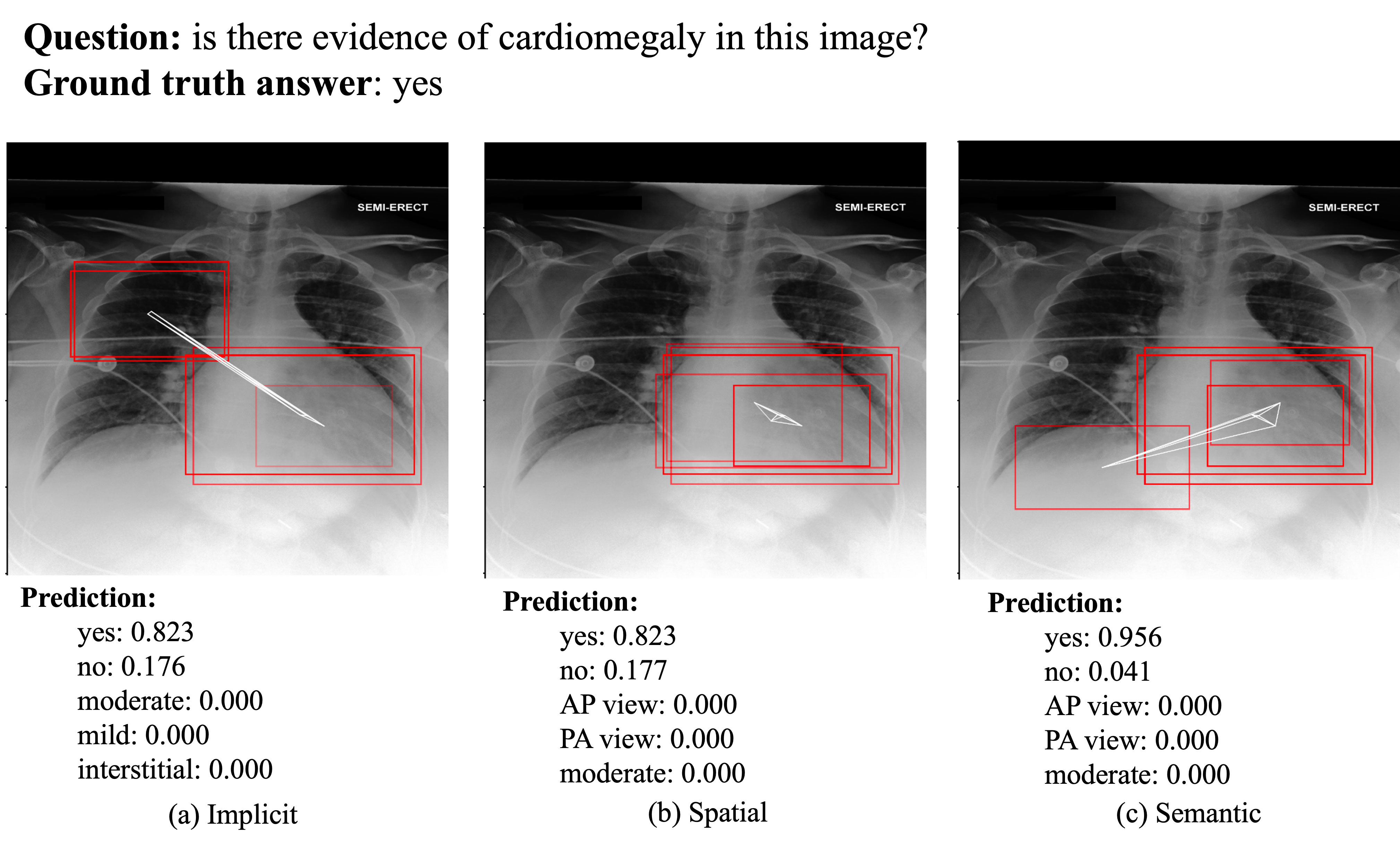}
    \caption{An example of the ROIs visualization for \textit{presence}. The red bounding boxes are the activated ROIs. 
    }
    \label{fig:visual}
\end{figure*}
\begin{figure*}[h]
    \centering
    \includegraphics[width=0.75\textwidth]{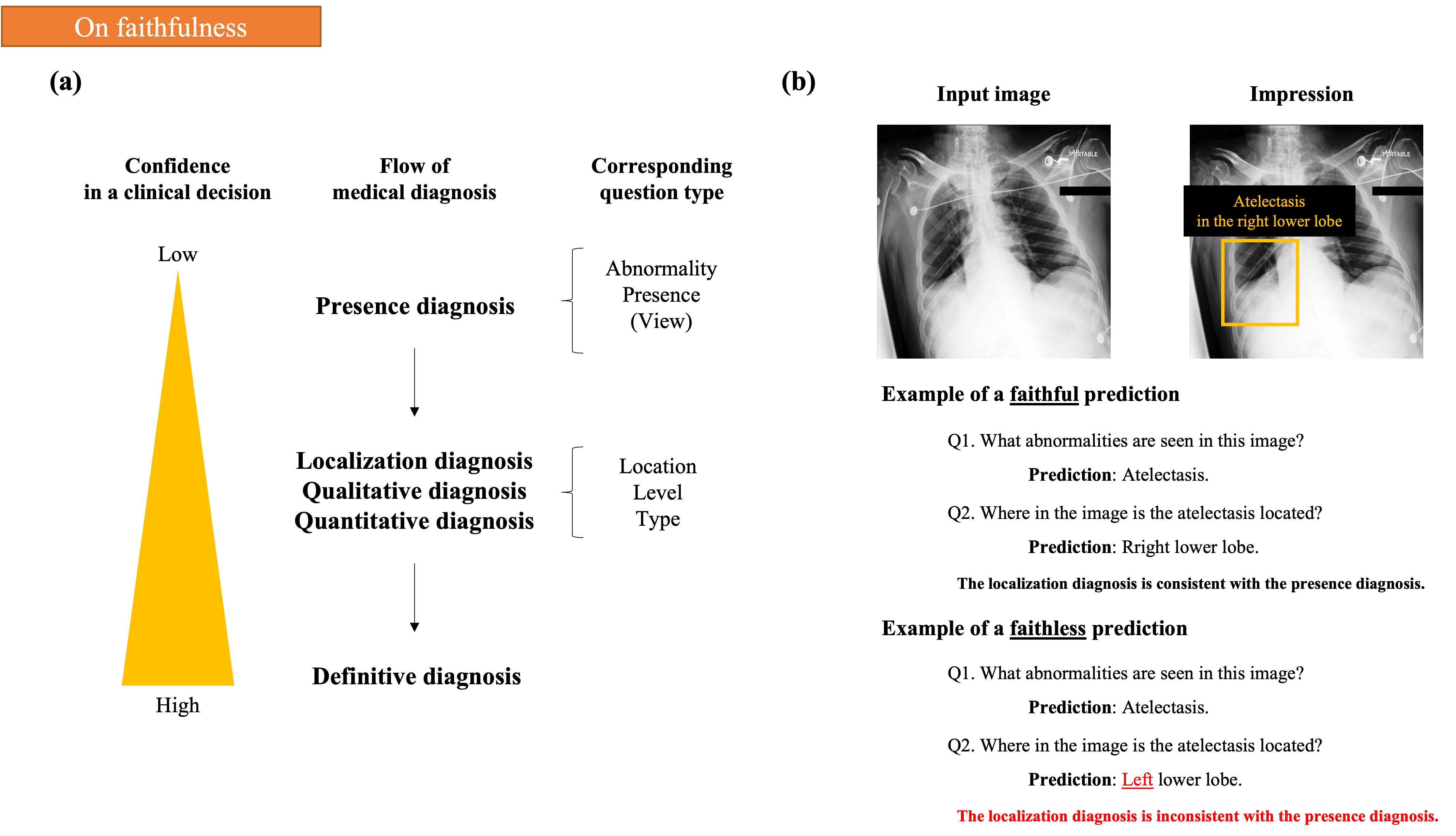}
    \caption{Illustration of faithfulness: (a) Increase in diagnosis confidence as finer questions are asked. (b) Examples of Faithful and Faithless Predictions. }
    \label{fig:faithfulness}
\end{figure*}
\begin{figure}[h]
    \centering
    \includegraphics[width=0.45\textwidth]{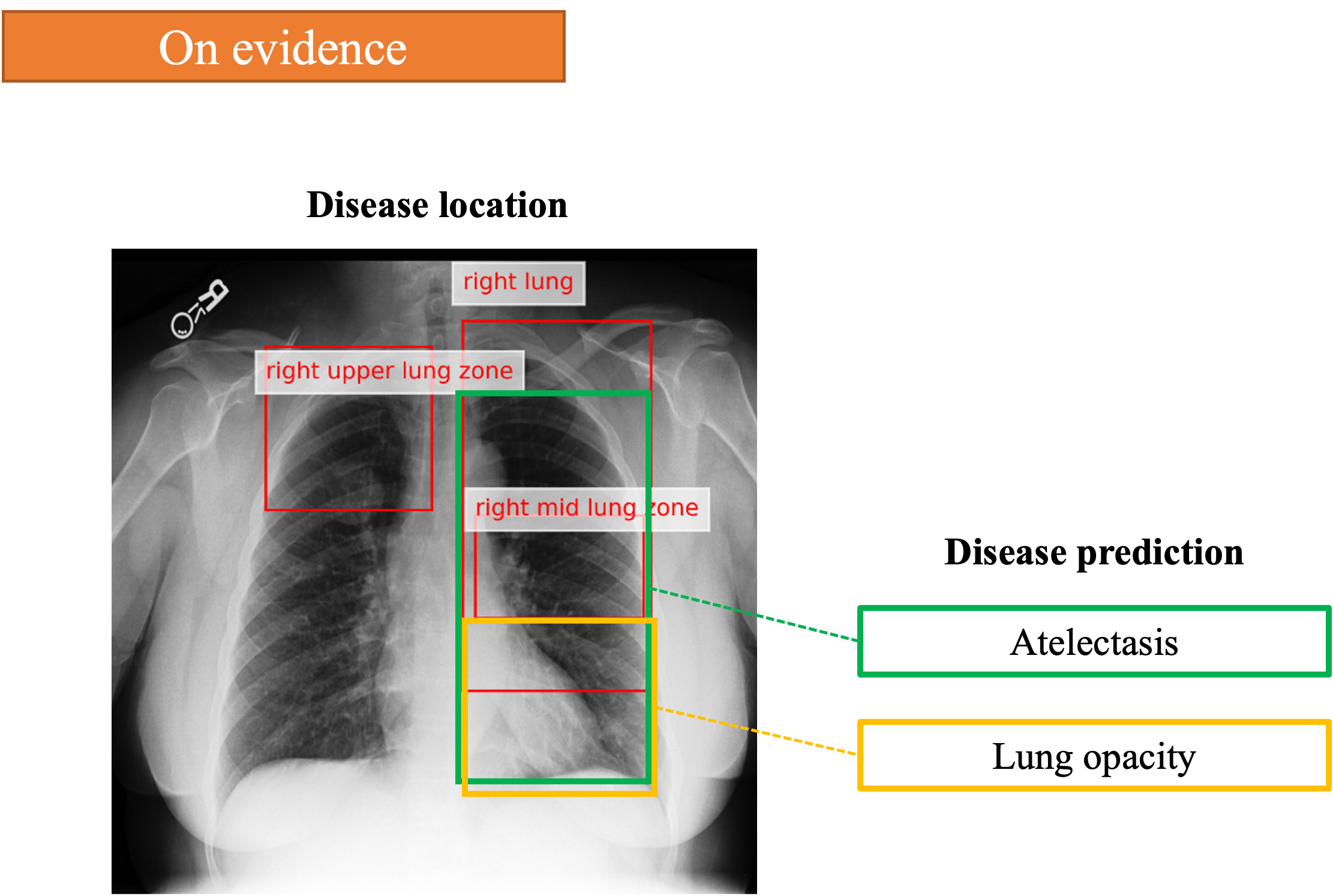}
    \caption{Illustration of evidence}
    \label{fig:evidence}
\end{figure}

\textbf{Visualization results.}
Here we present the learned relationships and ROIs to interpret the VQA answers. As shown in Fig.~\ref{fig:visual}, the input question in this example is "Is there any evidence of cardiomegaly in this image?", whose ground-truth answer is "yes".
We can see that the ROIs of all implicit, spatial, and semantic relationship graphs focus on the heart area, which is correct because cardiomegaly indicates an enlarged heart.
Furthermore, from the scores of each answer, we can see that our model successfully identifies this question as a closed question, \textit{i.e.}, a question with only "yes" or "no" as its possible answers.

\textbf{Please refer to the Appendix for more visualization examples of the other question types} including \textit{abnormality}, \textit{location}, \textit{type}, \textit{level}, and \textit{view} questions.

\section{Discussion.}
In the clinical field, it is crucial for an artificial intelligence tool to have both evidence and faithfulness. In this section, we will demonstrate that our method can provide both of these qualities.
As shown in Fig.~\ref{fig:evidence}, our method not only highlights the regions that are critical for predicting the final abnormalities, but it also provides location information for the corresponding abnormality by asking our model a location question. This provides the necessary evidence for doctors to inspect the diagnosis process.

In terms of faithfulness, as shown in Fig.~\ref{fig:faithfulness}(a), the medical diagnosis of a disease generally undergoes a course-to-fine fashion,  starting with a diagnosis of its presence in an organ or body (presence diagnosis) and progressing to precise localization and ultimately leading to a definitive diagnosis. In this process, the course level of diagnosis can be retrospectively validated by the finer level of information. When the finer information is consistent with the course diagnosis, the clinical decision can be made with confidence. Our question types can be classified into two groups: one corresponding to the presence diagnosis (e.g. abnormality, presence), and the other to the finer diagnosis (e.g. location, level) that relates to location, qualitative, and quantitative aspects. Therefore, by asking different groups of questions during the diagnosis, faithfulness can be achieved.

For example, in Fig.~\ref{fig:faithfulness}(b), our VQA model can provide clinical doctors with the opportunity to evaluate the faithfulness of the model prediction. Here, we consider a case with an input chest X-ray image with a doctor's impression of possible atelectasis in the left lower lobe. If the doctor asks the model if there is any abnormality in the image, and the model predicts the presence of atelectasis, we can further assess the accuracy of this prediction by asking the model for more specific information, such as the location of the atelectasis. If the model's localization diagnosis matches the doctor's impression, we can consider the model should comprehend the given clinical context. In contrast, if the localization diagnosis is inconsistent, the model prediction should not be trusted because it might overlook the actual pathology in the image. 

\textbf{Limitations.}
Although our method has demonstrated impressive performance, it is not without limitations. Our method may sometimes result in errors, including the following three: 1, confusion between different presentation aspects of the same abnormality, such as atelectasis and lung opacity being mistaken for each other. 2, different names for the same type of abnormality, such as enlargement of the cardiac silhouette being misclassified as cardiomegaly. 3, the pre-trained backbone (Faster-RCNN) used for extracting image features may provide inaccurate features and lead to incorrect predictions, such as lung opacity being wrongly recognized for pleural effusion. 

\section{Conclusion}
We compiled a large-scale and complicated medical VQA dataset, focusing on chest X-ray images. We also proposed a novel medical VQA baseline method based on multi-relationship graphs to incorporate spatial, semantic, and implicit relationships.
This method utilizes two types of knowledge graphs (anatomical and co-occurrence knowledge graphs) to model semantic relationships in medical visual question-answering tasks. We achieved a significant performance improvement compared to the state-of-the-art medical VQA methods.

\bibliographystyle{acl_natbib}
\bibliography{custom}

\section{Appendix}
\label{sec:appendix}

\subsection{Multi-Modal Graph Reasoning.}
We update each graph using Relation-Aware Graph Attention Network(ReGAT)~\cite{regat}.
When updating the graph, each neighbor node is multiplied with attention weights $\alpha_{ij}$ and a projection matrix $W$, where $i$ and $j$ represent the index of the node. 

\textbf{Implicit Graph Reasoning.}
Specifically, for the implicit relationship, the attention weights $\alpha_{ij}$ can be calcualted as below:
\begin{equation}
    \alpha_{ij} = \frac{\alpha_{ij}^b \cdot \exp{(\alpha_{ij}^v)}}{\sum_{j=1}^K \alpha_{ij}^b \cdot \exp{(\alpha_{ij}^v)}}
\end{equation}
\begin{equation}
    \alpha_{ij}^v = (\mathbf{U}\mathbf{v}_i)^\top \cdot (\mathbf{H}\mathbf{v}_j)
\end{equation}
\begin{equation}
    \alpha_{ij}^b = \max{(0, w\cdot f_b(\mathbf{b}_{ij}))}
\end{equation}
where $U$ and $H$ are projection matrix, $w$ is a transformation vector, $K$ is the number of the neighbor nodes, $\mathbf{b}_{ij}$ is the relative geometry feature between node $i$ and $j$, and can be calculated by $ [\log(\frac{|x_i-x_j|}{w_i}), \log(\frac{|y_i-y_j|}{h_i}), \log(\frac{w_j}{w_i}), \log(\frac{h_j}{h_i})]$, where  $x_i, x_j, y_i, y_j, w_i, w_j, h_i,$ and $h_j$. are the coordinates, width, and heights of the corresponding bounding box of the  node $i$, $f_b$ is a function that embeds the $4$-dimensional relative geometry feature into $d$-dimensional. 

Then, each updated node $\widetilde{\textbf{v}}_i \in \mathbb{R}^d$ in the final graph can be calculated as below:
\begin{equation}
    \widetilde{\textbf{v}}_i = \mathbf{W}^o \cdot (\|_{m=1}^M \sigma(\sum_{j\in \mathcal{N}_i} \alpha_{ij} \mathbf{W}^m \mathbf{v}_j))
\end{equation}
where $\mathcal{N}_i$ is the neighborhood set of the node $i$, $\mathbf{W}^m \in \mathbb{R}^{d \times (d_f + d_q)}$ is the projection matrix, $d$ is the dimension of the final node feature, $\sigma$ is the activation function, $\|_{m=1}^M$ represents concatenating the output of the $M$ attention heads, $\mathbf{W}^o \in \mathbb{R}^{d \times Md}$. 

\textbf{Spatial and Semantic Graph Reasoning}
For spatial and semantic graphs, which can also be called explicit graphs, can be seen as directed graphs. Therefore, the calculation of the attention weights and the updating of the graph consider the direction between node pairs and the labels of the edges. The attention weights can be calculated as follows:
\begin{equation}
    \alpha_{ij} = \frac{\exp{((\mathbf{U}\mathbf{v}_i)^\top \cdot \mathbf{H}_{dir(i,j)} \mathbf{v}_j + c_{lab(i,j)})}}{\sum_{j\in \mathcal{N}_i}\exp{((\mathbf{U}\mathbf{v}_i)^\top \cdot \mathbf{H}_{dir(i,j)} \mathbf{v}_j + c_{lab(i,j)})}}
\end{equation}
where $W_{dir(i,j)}, V_{dir(i,j)}\in  \mathbb{R}^{d \times (d_f + d_q)}$ are projection matrices, $b_{lab(i,j)}, c_{lab(i,j)}\in  \mathbb{R}^{d}$ are bias terms, $dir(i,j)$ represents the direction goes from node $i$ to $j$.

Then, the updated node $\widetilde{\textbf{v}}_i \in \mathbb{R}^d$ in the final graph can be calculated as:
\begin{equation}
    \widetilde{\textbf{v}}_i = \sigma(\sum_{j\in \mathcal{N}_i} \alpha_{ij} \mathbf{W}_{dir(i,j)} \mathbf{v}_j + b_{lab(i,j)})
\end{equation}
where $lab(i,j)$ represents the label assigned to the edge $(i,j)$.

Similarly, the multi-head attention can be calculated by concatenating the output features and adding a projection matrix $\mathbf{W}^o \in \mathbb{R}^{d \times Md}$.

\textbf{Final Feature Vector}
Finally, the feature vector $a \in \mathbb{R}^{c}$ of one relationship graph is calculated by 
\begin{equation}
    a = f(\widetilde{V}),
\end{equation}
where $c$ is the number of the classes, $f(\cdot)$ is the multi-layer perceptron. 
For the final feature vector, $a_{final}$ can be calculated by:
\begin{equation}
    a_{final} = (1 - \alpha - \beta) \times a_{imp} + \alpha \times a_{spa} + \beta \times a_{sem}
\end{equation}
where $a_{imp}, a_{spa}, a_{sem}$ are the feature vector of the implicit graph, spatial graph, and semantic graph, respectively, and $\alpha, \beta$ are coefficients.

\subsection{Visualizations}
Fig.~\ref{fig:location} demonstrates the visualization of a \textit{loacation} question. The question asks "Is the opacity located on the left side or right side". The ground truth is "right side". Very intuitively, all ROIs are focusing on the right lung area (The right side of the patient is the left side of the picture). 

Fig.~\ref{fig:abnormality} is an example of the visualization of an \textit{abnormality} question. The ground truth answer to this question is "cardiomegaly, pleural effusion, atelectasis, lung opacity", which covers both the lung and heart regions. We can observe that these regions are attended to in all three relation graphs.

Fig.~\ref{fig:level} shows a visualization of \textit{level} question. In this example, mild edema is observed in both lungs according to the corresponding medical reports. The ROIs are activated in both lungs.

Lastly, we have another example of a \textit{view} question, which is shown in Fig.~\ref{fig:view}. PA view and AP view can be differentiated by the direction of the ribs and the contour of the heart. PA view typically has a more slender heart shape. The ROIs on the rib area and heart area are activated.

\newpage
\begin{figure*}[t]
   \centering
       \begin{subfigure}[b]{\textwidth}
       \centering
       \includegraphics[width=\textwidth]{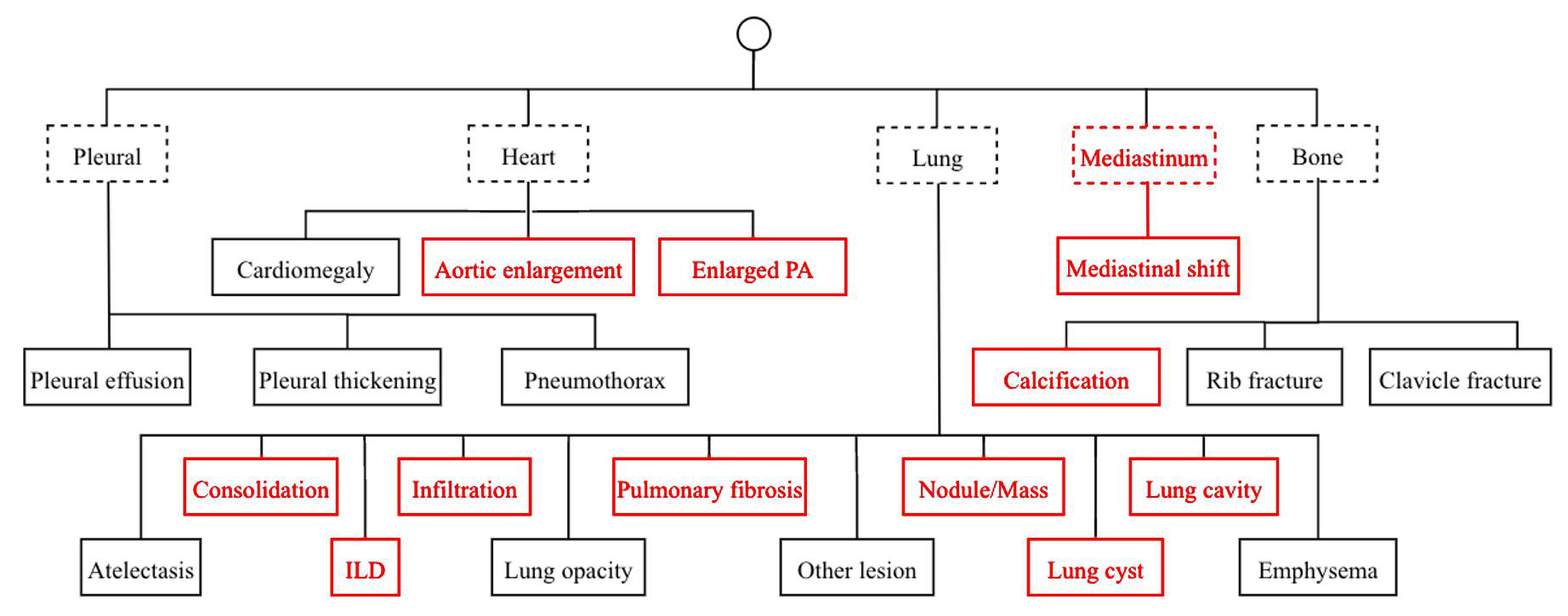}
       \caption{Anatomical Knowledge Graph}
       \label{fig:anaKG}
       \end{subfigure}
        
       \begin{subfigure}[b]{\textwidth}
       \centering
       \includegraphics[width=0.8\textwidth]{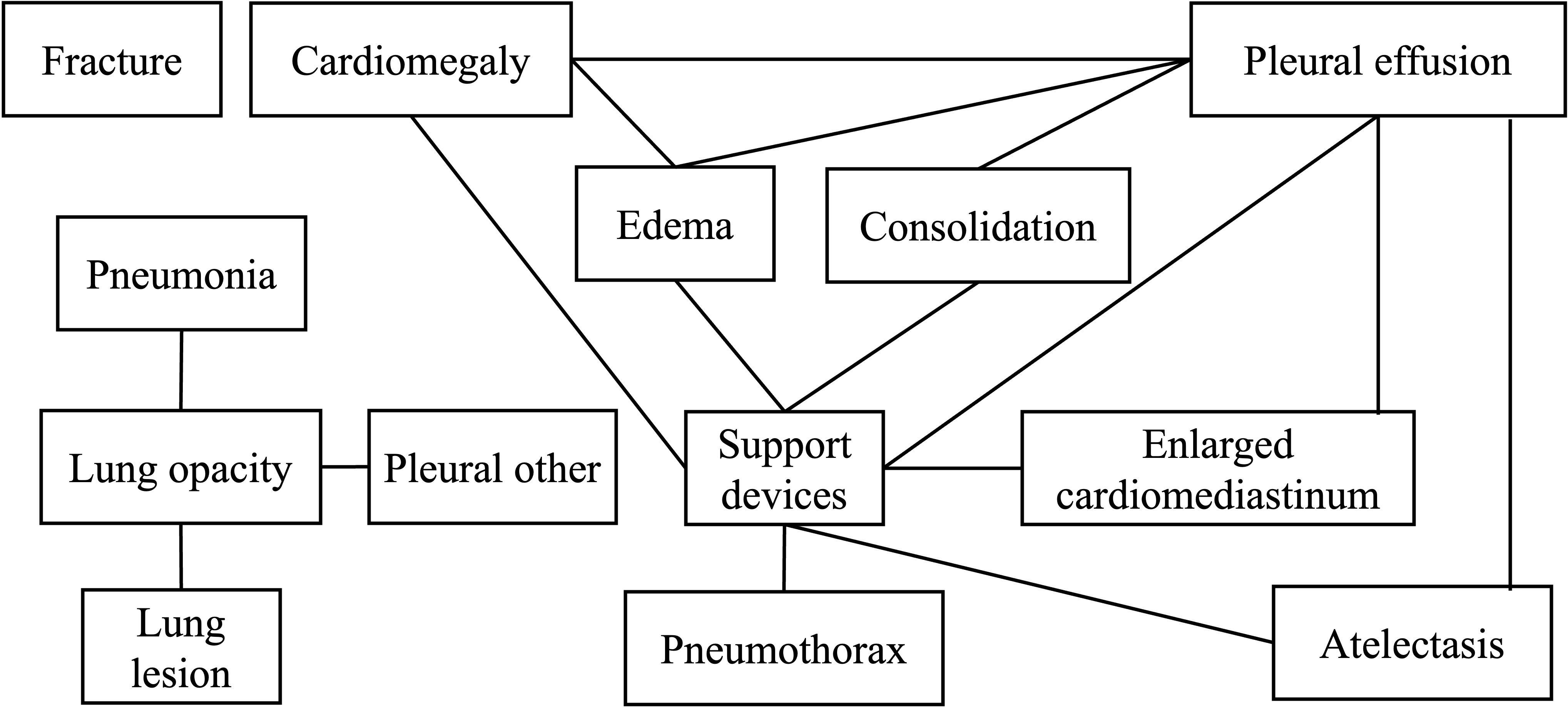}
        \caption{Co-occurrence Knowledge Graph}
        \label{fig:coKG}
        \end{subfigure}
        
        
   \caption{Knowledge Graphs.}
   \label{fig:KG}
\end{figure*}

\begin{table*}[t]
\caption{Abnormality Keywords}
\label{tab:dis}
\centering
\resizebox{2\columnwidth}{!}{%
\begin{tabular}{ll}
\toprule
id & Abnormality names                                                                                                            \\
\midrule
0  & pleural effusions;pleural effusion;effusion;effusions;pleural fluid                                                                                                                                                                            \\
1  & volume loss;collapse;atelectasis;atelectases;atelectatic   changes;atelectatic change                                                                                                                                                          \\
2  & cardiomegaly;heart size is enlarged                                                                                                                                                                                                            \\
3  & enlargement of the cardiac silhouette                                                                                                                                                                                                          \\
4  & pulmonary edema;edema                                                                                                                                                                                                                          \\
5  & hiatal hernia;hiatus hernia;hernia                                                                                                                                                                                                             \\
6  & pulmonary vascular congestion;vascular congestion                                                                                                                                                                                              \\
7  & hilar congestion                                                                                                                                                                                                                               \\
8  & pneumothorax                                                                                                                                                                                                                                   \\
9  & cardiac decompensation;chf;congestive heart failure;heart failure                                                                                                                                                                              \\
10 & lung opacification;airspace opacities;airspace   opacity;opacification;opacity;opacities;lung opacity;lung opacities                                                                                                                           \\
11 & pneumonia;infection                                                                                                                                                                                                                            \\
12 & tortuosity of the descending aorta                                                                                                                                                                                                             \\
13 & thoracolumbar scoliosis;scoliosis                                                                                                                                                                                                              \\
14 & gastric distention                                                                                                                                                                                                                             \\
15 & hypoxemia                                                                                                                                                                                                                                      \\
16 & hypertensive heart disease;htn                                                                                                                                                                                                                 \\
17 & hematoma                                                                                                                                                                                                                                       \\
18 & tortuosity of the thoracic aorta                                                                                                                                                                                                               \\
19 & pulmonary contusion;contusion                                                                                                                                                                                                                  \\
20 & emphysema                                                                                                                                                                                                                                      \\
21 & granulomatous disease;granuloma                                                                                                                                                                                                                \\
22 & calcifications;calcification                                                                                                                                                                                                                   \\
23 & pleural thickening                                                                                                                                                                                                                             \\
24 & thymoma                                                                                                                                                                                                                                        \\
25 & blunting of the costophrenic angles;blunting of the right costophrenic   angle;blunting of the left costophrenic angle;\\
  & blunting of the costophrenic   angle;blunting of the left costodiaphragmatic;blunting of the right   costodiaphragmatic \\
26 & consolidation                                                                                                                                                                                                                                  \\
27 & fractures;fracture                                                                                                                                                                                                                             \\
28 & pneumomediastinum                                                                                                                                                                                                                              \\
29 & air collection                                       \\
\bottomrule
\end{tabular}%
}
\end{table*}

\begin{table*}[h]
\caption{Attribute keywords for level, location(pre), location(post), and type.}
\label{tab:attr}
\centering
\begin{tabular}{llll}
\toprule
\multicolumn{4}{c}{Attribute}     
\\
\midrule
level              & location(pre) & location(post)                    & type         \\
\hline
moderate           & mid to lower  & the lower lobe                    & interstitial \\
acute              & left          & the upper lobe                    & layering     \\
mild               & right         & the middle lobe                   & dense        \\
small              & retrocardiac  & the left lung base                & parenchymal  \\
moderately         & pericardial   & the right lung base               & compressive  \\
severe             & bibasilar     & the lung bases                    & obstructive  \\
moderate to large  & bilateral     & the left base                     & linear       \\
moderate to severe & basilar       & the right base                    & plate-like   \\
mild to moderate   & apicolateral  & the right upper lung              & patchy       \\
moderate to large  & basal         & the left upper lung               & ground-glass \\
minimal            & left-sided    & the right middle lung             & calcified    \\
mildly             & lobe          & the left middle lung              & scattered    \\
subtle             & lung          & the right mid lung                & interstitial \\
                  & area          & the left mid lung                 & focal        \\
                  & right-sided   & the right lower lung              & multifocal   \\
                  & apical        & the left lower lung               & multi-focal  \\
                  & pleural       & the right upper lobe              &              \\
                  & upper         & the left upper lobe               &              \\
                  & lower         & the right middle lobe             &              \\
                  & middle        & the left middle lobe              &              \\
                  & mid           & the right mid lobe                &              \\
                  & rib           & the left mid lobe                 &              \\
                  &               & the right lower lobe              &              \\
                  &               & the left lower lobe               &              \\
                  &               & the left apical area              &              \\
                  &               & the left apical region            &              \\
                  &               & the right apical area             &              \\
                  &               & the right apical region           &              \\
                  &               & the apical region                 &              \\
                  &               & the apical area                   &              \\
                  &               & the right mid to lower lung       &              \\
                  &               & the left mid to lower lung        &              \\
                  &               & the medial right lung base        &              \\
                  &               & the medial left lung base         &              \\
                  &               & the upper lungs                   &              \\
                  &               & the lower lungs                   &              \\
                  &               & the upper lobes                   &              \\
                  &               & the lower lobes                   &              \\
                  &               & the right mid to lower hemithorax &              \\
                  &               & the left mid to lower hemithorax  &   \\  
\bottomrule
\end{tabular}
\end{table*}

\begin{figure*}[h]
    \centering
    \includegraphics[width=0.645\textwidth]{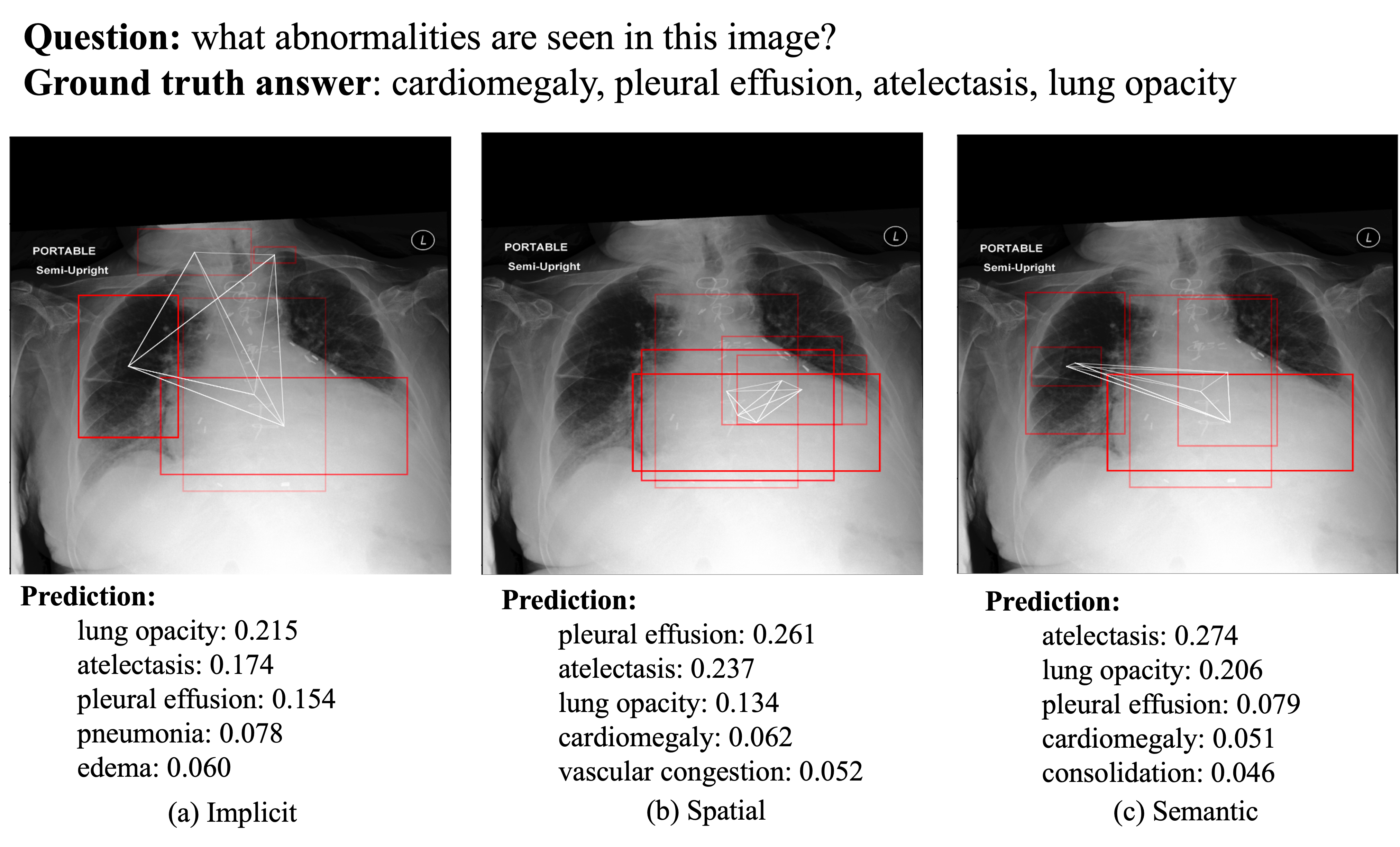}
    \caption{An example of the ROIs visualization for \textit{abnormality}. The red bounding boxes are the activated ROIs. 
    }
    \label{fig:abnormality}
\end{figure*}
\begin{figure*}[h]
    \centering
    \includegraphics[width=0.645\textwidth]{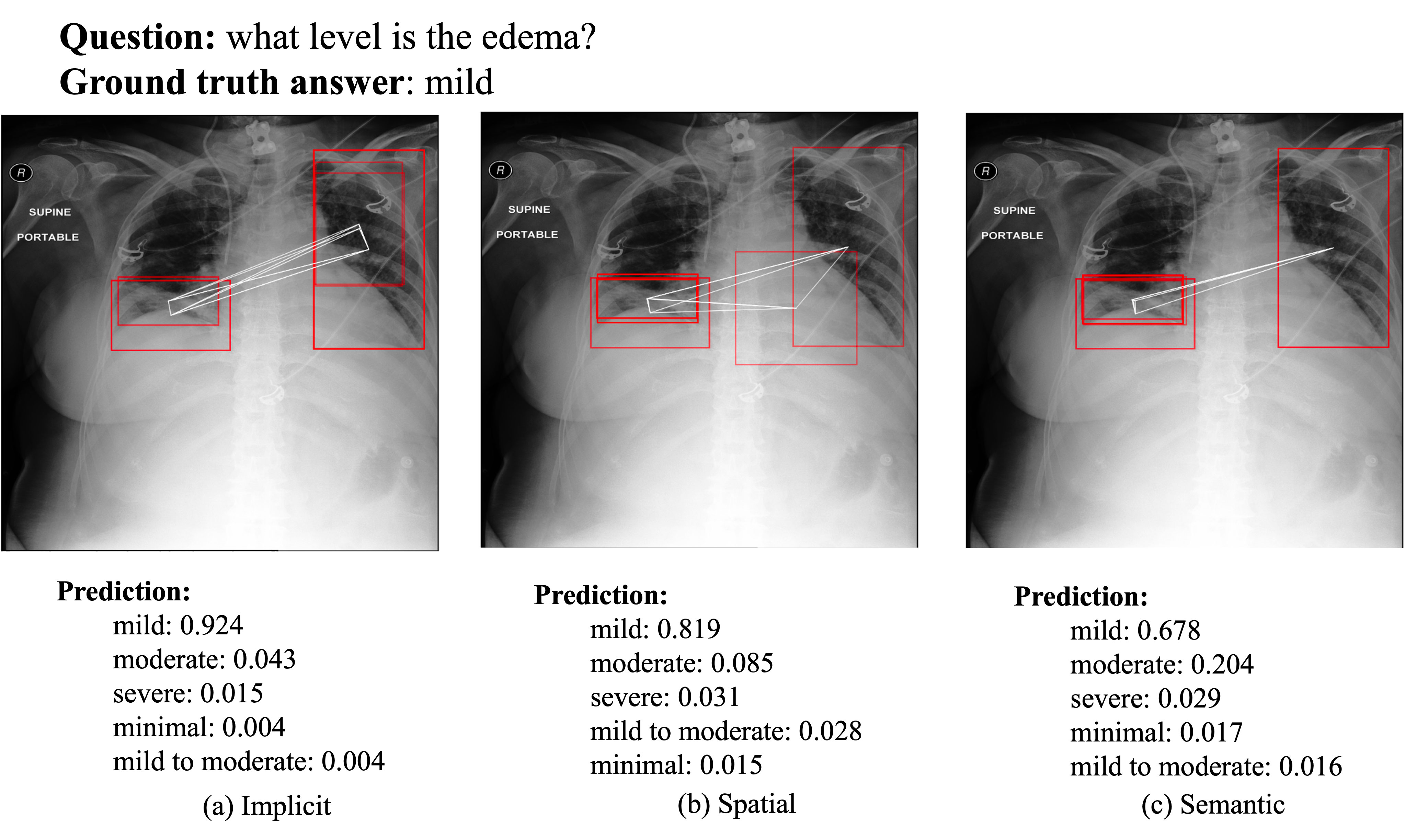}
    \caption{An example of the ROIs visualization for \textit{level}. The red bounding boxes are the activated ROIs. 
    }
    \label{fig:level}
\end{figure*}
\begin{figure*}[t]
    \centering
    \includegraphics[width=0.66\textwidth]{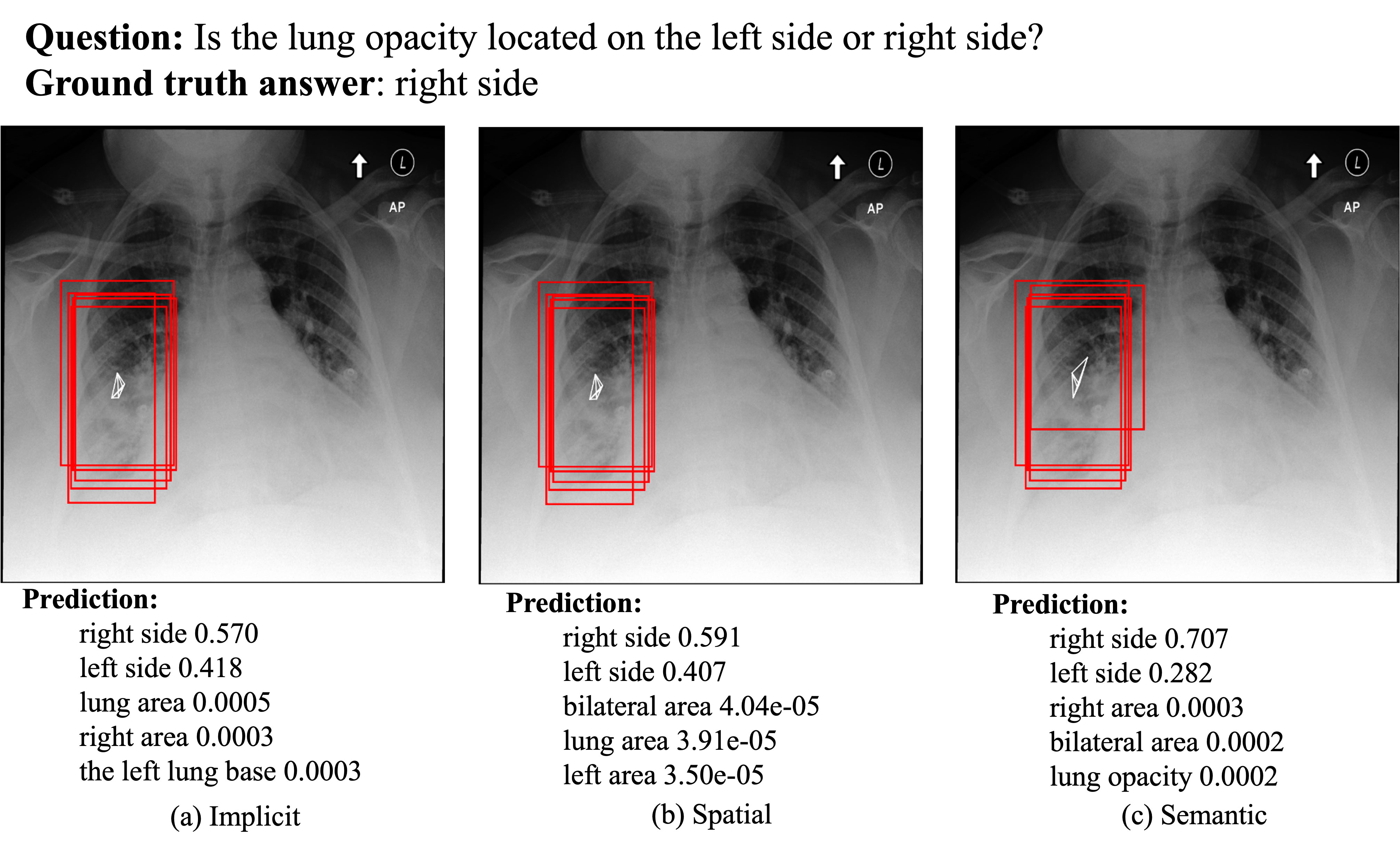}
    \caption{An example of the visualization result for \textit{location}. The red bounding boxes are the activated ROIs. 
    }
    \label{fig:location}
\end{figure*}
\begin{figure*}[t]
    \centering
    \includegraphics[width=0.8\textwidth]{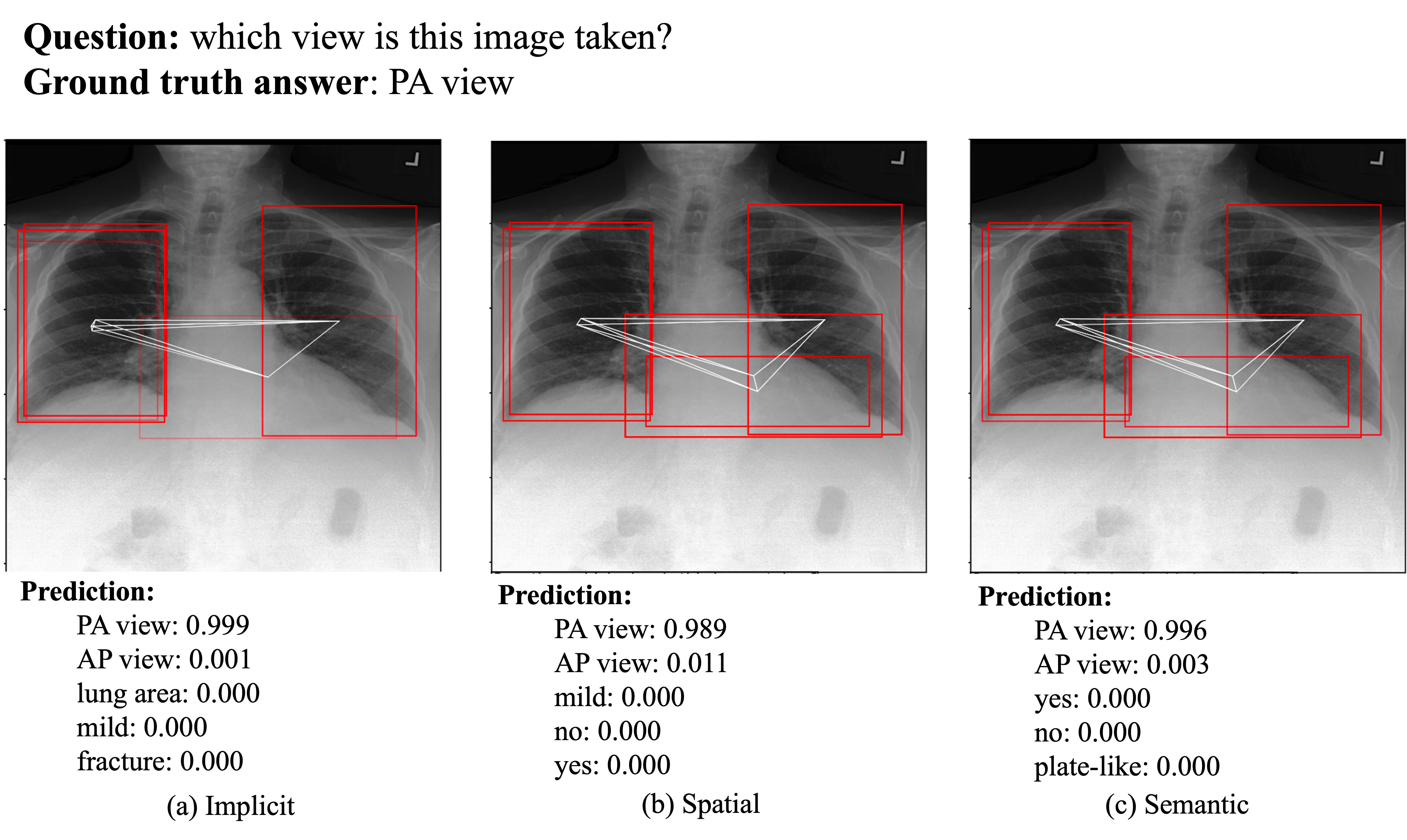}
    \caption{An example of the ROIs visualization for \textit{view}. The red bounding boxes are the activated ROIs. 
    }
    \label{fig:view}
\end{figure*}

\newpage
\clearpage
\onecolumn
\begin{longtable}[c]{l|l|llll}
\caption{The quantitative results for each label. "imp", "spa", "sem" and "com" represent "implicit", "spatial", "semantic", and "combined", respectively.}
\label{tab:longtab}\\
\multirow{2}{*}{Labels}               & \multirow{2}{*}{MMQ} & \multicolumn{4}{c}{Ours}                                          \\ \cline{3-6}
                                      &                      & imp      & spa        & sem       & com       \\ \hline
\endfirsthead
\multicolumn{6}{c}%
{{\bfseries Table \thetable\ continued from previous page}} \\
\multirow{2}{*}{Labels}               & \multirow{2}{*}{MMQ} & \multicolumn{4}{c}{Ours}                                          \\ \cline{3-6} 
                                      &                      & imp      & spa        & sem       & com       \\ \hline
\endhead
AP view                               & 0.987                & 0.986          & 0.986          & 0.991          & \textbf{0.989} \\ \hline
PA view                               & \textbf{1.000}       & 0.998          & 0.998          & 0.999          & 0.999          \\
acute                                 & 0.871                & 0.977          & 0.965          & 0.980          & \textbf{0.981} \\
apical area                           & 0.930                & \textbf{0.974} & \textbf{0.974} & 0.964          & 0.972          \\
apical right area                     & 0.356                & 0.582          & 0.696          & \textbf{0.857} & 0.735          \\
area area                             & 0.907                & 0.960          & 0.873          & \textbf{0.961} & 0.957          \\
atelectasis                           & 0.956                & 0.964          & \textbf{0.966} & 0.958          & 0.964          \\
basal area                            & 0.909                & 0.939          & 0.930          & \textbf{0.946} & \textbf{0.949} \\
basilar area                          & 0.888                & 0.941          & 0.939          & \textbf{0.951} & 0.948          \\
basilar lung area                     & 0.714                & \textbf{0.974} & 0.896          & 0.877          & 0.924          \\
bibasilar area                        & 0.933                & 0.948          & 0.944          & 0.950          & \textbf{0.951} \\
bibasilar retrocardiac area           & 0.950                & \textbf{0.878} & 0.875          & 0.877          & 0.874          \\
bilateral apical area                 & 0.960                & \textbf{0.968} & 0.917          & 0.709          & 0.889          \\
bilateral area                        & 0.913                & 0.969          & \textbf{0.970} & \textbf{0.970} & \textbf{0.970} \\
bilateral basal area                  & 0.848                & 0.912          & 0.911          & \textbf{0.921} & 0.918          \\
bilateral basilar area                & 0.876                & 0.936          & 0.886          & \textbf{0.958} & 0.935          \\
bilateral lower lung area             & 0.914                & 0.913          & 0.923          & \textbf{0.930} & 0.922          \\
bilateral lung area                   & 0.802                & \textbf{0.959} & 0.934          & 0.950          & 0.953          \\
bilateral retrocardiac area           & 0.610                & \textbf{0.976} & 0.936          & 0.936          & 0.967          \\
bilateral rib area                    & 0.978                & \textbf{0.996} & 0.995          & \textbf{0.996} & \textbf{0.996} \\
bilateral upper lung area             & 0.856                & 0.829          & 0.854          & \textbf{0.871} & 0.870          \\
blunting of the costophrenic angle    & 0.923                & 0.948          & 0.952          & \textbf{0.953} & \textbf{0.953} \\
calcification                         & 0.910                & \textbf{0.945} & 0.946          & 0.942          & \textbf{0.945} \\
calcified                             & 0.996                & \textbf{0.999} & \textbf{0.999} & 0.998          & \textbf{0.999} \\
calcified calcified                   & 0.647                & \textbf{0.980} & 0.928          & 0.739          & 0.943          \\
cardiomegaly                          & \textbf{0.948}       & 0.947          & 0.943          & 0.943          & 0.945          \\
compressive                           & 0.985                & \textbf{0.999} & \textbf{0.999} & \textbf{0.999} & \textbf{0.999} \\
consolidation                         & 0.919                & 0.934          & \textbf{0.938} & 0.933          & 0.936          \\
contusion                             & 0.834                & \textbf{0.927} & 0.863          & 0.917          & 0.913          \\
dense                                 & 0.946                & \textbf{0.989} & 0.986          & 0.986          & \textbf{0.989} \\
edema                                 & 0.952                & \textbf{0.970} & 0.968          & 0.965          & 0.969          \\
emphysema                             & 0.920                & 0.944          & \textbf{0.946} & 0.945          & \textbf{0.946} \\
enlargement of the cardiac silhouette & 0.936                & \textbf{0.942} & 0.919          & 0.935          & 0.936          \\
focal                                 & 0.927                & 0.985          & \textbf{0.988} & 0.986          & \textbf{0.988} \\
focal parenchymal                     & 0.859                & \textbf{0.983} & 0.980          & 0.961          & 0.979          \\
focal patchy                          & 0.876                & \textbf{0.985} & 0.977          & 0.961          & 0.978          \\
fracture                              & 0.922                & 0.940          & 0.940          & \textbf{0.942} & 0.941          \\
gastric distention                    & 0.724                & 0.910          & 0.814          & \textbf{0.917} & 0.906          \\
granuloma                             & 0.940                & \textbf{0.965} & 0.962          & 0.948          & 0.961          \\
ground-glass                          & 0.932                & \textbf{0.988} & 0.983          & 0.985          & 0.987          \\
heart failure                         & 0.891                & \textbf{0.948} & 0.936          & 0.936          & 0.942          \\
hematoma                              & 0.904                & 0.909          & \textbf{0.915} & 0.910          & 0.908          \\
hernia                                & 0.918                & 0.942          & 0.939          & \textbf{0.947} & 0.944          \\
hilar congestion                      & 0.824                & \textbf{0.944} & 0.927          & 0.932          & 0.942          \\
infection                             & 0.919                & 0.931          & \textbf{0.939} & 0.933          & 0.938          \\
interstitial                          & 0.974                & \textbf{0.996} & \textbf{0.996} & 0.994          & \textbf{0.996} \\
interstitial parenchymal              & 0.665                & \textbf{0.991} & 0.959          & 0.919          & 0.981          \\
layering                              & \textbf{1.000}       & 0.997          & 0.999          & 0.997          & 0.997          \\
left apical area                      & 0.912                & 0.985          & \textbf{0.987} & 0.985          & \textbf{0.987} \\
left area                             & 0.889                & \textbf{0.957} & 0.955          & 0.956          & \textbf{0.957} \\
left basal area                       & 0.913                & 0.936          & 0.931          & 0.937          & \textbf{0.940} \\
left basal retrocardiac area          & 0.874                & 0.737          & \textbf{0.816} & 0.616          & 0.705          \\
left basilar area                     & 0.925                & 0.943          & \textbf{0.944} & 0.943          & \textbf{0.944} \\
left basilar retrocardiac area        & 0.838                & 0.938          & 0.916          & 0.944          & \textbf{0.945} \\
left bibasilar area                   & 0.724                & 0.756          & \textbf{0.942} & 0.895          & 0.906          \\
left lower area                       & 0.831                & 0.926          & 0.922          & \textbf{0.945} & 0.936          \\
left lower lung area                  & 0.926                & \textbf{0.949} & 0.945          & 0.948          & \textbf{0.949} \\
left lower lung retrocardiac area     & 0.760                & 0.922          & \textbf{0.926} & 0.899          & 0.919          \\
left lower rib area                   & 0.201                & 0.991          & 0.988          & 0.988          & \textbf{0.992} \\
left lung area                        & 0.917                & 0.926          & 0.935          & \textbf{0.944} & 0.941          \\
left mid area                         & \textbf{0.965}       & 0.951          & 0.860          & 0.929          & 0.934          \\
left middle lung area                 & 0.871                & 0.943          & \textbf{0.948} & 0.928          & 0.946          \\
left pleural area                     & 0.962                & \textbf{1.000} & 0.997          & 0.874          & 0.996          \\
left retrocardiac area                & 0.938                & 0.950          & 0.950          & \textbf{0.955} & 0.953          \\
left rib area                         & 0.975                & \textbf{0.996} & \textbf{0.996} & \textbf{0.996} & \textbf{0.996} \\
left right area                       & 0.735                & 0.954          & \textbf{0.961} & 0.958          & \textbf{0.961} \\
left right basal area                 & 0.767                & 0.856          & 0.857          & \textbf{0.931} & 0.903          \\
left right bibasilar area             & 0.792                & 0.744          & 0.863          & \textbf{0.901} & 0.866          \\
left side                             & 0.956                & 0.980          & 0.981          & \textbf{0.982} & \textbf{0.982} \\
left upper area                       & 0.772                & 0.924          & 0.934          & 0.937          & \textbf{0.954} \\
left upper lung area                  & 0.906                & 0.927          & 0.938          & \textbf{0.945} & 0.941          \\
linear                                & 0.966                & \textbf{0.991} & \textbf{0.991} & \textbf{0.991} & \textbf{0.991} \\
linear patchy                         & 0.828                & 0.844          & \textbf{0.937} & 0.785          & 0.914          \\
lower area                            & 0.897                & 0.921          & 0.938          & \textbf{0.941} & 0.940          \\
lower lung area                       & 0.910                & 0.930          & 0.928          & \textbf{0.943} & 0.940          \\
lung area                             & \textbf{0.935}       & \textbf{0.953} & 0.944          & 0.946          & \textbf{0.953} \\
lung basilar area                     & 0.889                & 0.926          & 0.956          & \textbf{0.979} & 0.969          \\
lung bibasilar area                   & 0.890                & 0.946          & 0.943          & 0.940          & \textbf{0.948} \\
lung bilateral area                   & 0.514                & 0.903          & \textbf{0.945} & 0.834          & 0.909          \\
lung left area                        & 0.449                & \textbf{0.814} & 0.700          & 0.697          & 0.741          \\
lung opacity                          & 0.950                & \textbf{0.954} & 0.953          & 0.953          & \textbf{0.954} \\
lung right area                       & 0.414                & \textbf{0.632} & 0.548          & 0.390          & 0.518          \\
lung right middle lung area           & 0.672                & 0.794          & \textbf{0.685} & 0.600          & 0.672          \\
mid area                              & 0.812                & 0.954          & 0.973          & 0.973          & \textbf{0.976} \\
middle left lung area                 & \textbf{0.970}       & 0.921          & 0.845          & 0.963          & 0.948          \\
middle lower lung area                & \textbf{0.949}       & 0.930          & 0.915          & 0.929          & 0.933          \\
middle lung area                      & 0.941                & 0.951          & 0.955          & \textbf{0.987} & 0.978          \\
middle to lower left lung area        & 0.598                & 0.901          & \textbf{0.928} & 0.902          & 0.924          \\
middle to lower lung area             & \textbf{0.963}       & 0.447          & 0.751          & 0.918          & 0.671          \\
middle to lower right lung area       & \textbf{0.862}       & 0.859          & 0.767          & 0.716          & 0.787          \\
mild                                  & 0.935                & \textbf{0.977} & \textbf{0.977} & \textbf{0.977} & \textbf{0.977} \\
mild mild                             & 0.364                & 0.560          & 0.684          & \textbf{0.890} & 0.793          \\
mild moderate                         & 0.584                & 0.513          & 0.494          & \textbf{0.961} & 0.707          \\
mild to moderate                      & 0.765                & 0.956          & \textbf{0.957} & 0.956          & 0.959          \\
mildly                                & 0.860                & \textbf{0.938} & 0.896          & 0.934          & 0.931          \\
mildly mild                           & \textbf{0.276}       & 0.078          & 0.230          & 0.359          & 0.202          \\
minimal                               & 0.852                & 0.966          & 0.969          & 0.970          & \textbf{0.971} \\
minimal mild                          & 0.149                & 0.341          & 0.206          & \textbf{0.534} & 0.347          \\
minimal moderate                      & 0.400                & 0.176          & 0.376          & \textbf{0.480} & 0.321          \\
moderate                              & 0.858                & 0.961          & \textbf{0.963} & 0.961          & \textbf{0.963} \\
moderate moderately severe            & \textbf{0.445}       & 0.207          & 0.273          & 0.239          & 0.227          \\
moderate small                        & 0.749                & \textbf{0.972} & 0.965          & 0.965          & 0.971          \\
moderate to large                     & 0.922                & 0.963          & 0.963          & 0.961          & \textbf{0.964} \\
moderate to large moderate            & 0.363                & 0.333          & 0.389          & \textbf{0.457} & 0.379          \\
moderate to large small               & 0.641                & 0.616          & 0.438          & \textbf{0.621} & 0.498          \\
moderate to severe                    & 0.859                & 0.951          & 0.951          & \textbf{0.955} & \textbf{0.955} \\
moderately                            & 0.612                & \textbf{0.940} & 0.909          & 0.878          & 0.932          \\
moderately severe                     & 0.809                & 0.960          & 0.931          & \textbf{0.982} & 0.969          \\
multi-focal                           & \textbf{0.997}       & 0.510          & 0.783          & 0.904          & 0.708          \\
multifocal                            & 0.961                & \textbf{0.994} & \textbf{0.994} & 0.992          & \textbf{0.994} \\
multifocal parenchymal                & 0.765                & \textbf{0.986} & 0.983          & 0.979          & 0.985          \\
no                                    & 0.959                & \textbf{0.992} & 0.991          & 0.991          & \textbf{0.992} \\
obstructive                           & 0.832                & 0.946          & \textbf{0.988} & 0.938          & 0.977          \\
parenchymal                           & 0.939                & \textbf{0.992} & 0.991          & 0.991          & \textbf{0.992} \\
patchy                                & 0.927                & \textbf{0.985} & \textbf{0.985} & 0.983          & \textbf{0.985} \\
patchy linear                         & 0.931                & 0.975          & 0.976          & 0.979          & \textbf{0.980} \\
patchy parenchymal                    & 0.911                & 0.940          & \textbf{0.969} & 0.705          & 0.931          \\
pericardial area                      & 0.833                & \textbf{0.967} & 0.952          & 0.956          & 0.960          \\
plate-like                            & 0.960                & 0.994          & 0.994          & \textbf{0.995} & 0.994          \\
pleural area                          & 0.877                & 0.923          & \textbf{0.945} & 0.931          & 0.939          \\
pleural effusion                      & 0.969                & \textbf{0.979} & \textbf{0.979} & 0.976          & \textbf{0.979} \\
pleural left area                     & 0.646                & \textbf{0.951} & 0.858          & 0.893          & 0.926          \\
pleural right area                    & 0.837                & \textbf{0.978} & 0.808          & 0.811          & 0.904          \\
pleural thickening                    & 0.931                & \textbf{0.955} & 0.951          & 0.953          & \textbf{0.955} \\
pneumonia                             & 0.932                & 0.943          & \textbf{0.945} & 0.937          & 0.944          \\
pneumothorax                          & 0.921                & 0.941          & 0.942          & 0.937          & \textbf{0.943} \\
retrocardiac area                     & 0.927                & 0.942          & 0.945          & 0.944          & \textbf{0.947} \\
retrocardiac area area                & 0.827                & 0.771          & 0.911          & \textbf{0.917} & 0.904          \\
retrocardiac left lower lung area     & 0.792                & 0.821          & 0.903          & \textbf{0.972} & 0.938          \\
retrocardiac right basal area         & 0.777                & 0.896          & \textbf{0.953} & 0.829          & 0.922          \\
retrocardiac right basilar area       & 0.909                & 0.893          & 0.904          & \textbf{0.944} & 0.923          \\
rib area                              & 0.993                & \textbf{0.997} & \textbf{0.997} & \textbf{0.997} & \textbf{0.997} \\
right apical area                     & 0.963                & 0.990          & \textbf{0.993} & 0.992          & 0.992          \\
right area                            & 0.880                & \textbf{0.958} & 0.957          & 0.955          & \textbf{0.958} \\
right basal area                      & 0.881                & 0.921          & 0.926          & \textbf{0.931} & 0.928          \\
right basilar area                    & 0.906                & \textbf{0.942} & 0.930          & 0.937          & 0.939          \\
right bibasilar area                  & 0.580                & 0.853          & 0.867          & 0.881          & \textbf{0.889} \\
right left area                       & 0.763                & \textbf{0.970} & 0.967          & 0.963          & 0.969          \\
right left basilar area               & \textbf{0.953}       & 0.897          & 0.926          & 0.716          & 0.855          \\
right left rib area                   & \textbf{0.972}       & 0.955          & 0.810          & 0.960          & 0.952          \\
right lower area                      & 0.850                & \textbf{0.936} & 0.929          & 0.905          & 0.926          \\
right lower lung area                 & 0.929                & 0.946          & \textbf{0.947} & 0.937          & 0.944          \\
right lower middle lung area          & \textbf{0.912}       & 0.835          & 0.874          & 0.838          & 0.860          \\
right lower rib area                  & 0.997                & 0.995          & 0.996          & \textbf{0.999} & 0.997          \\
right lung area                       & 0.894                & 0.934          & \textbf{0.941} & 0.935          & 0.939          \\
right mid area                        & 0.890                & \textbf{0.949} & 0.920          & 0.880          & 0.923          \\
right middle lower area               & \textbf{0.983}       & 0.568          & 0.754          & 0.681          & 0.655          \\
right middle lower lung area          & \textbf{0.958}       & 0.946          & 0.888          & 0.907          & 0.919          \\
right middle lung area                & 0.930                & \textbf{0.948} & 0.942          & 0.933          & 0.945          \\
right pleural area                    & 0.667                & \textbf{0.849} & 0.802          & 0.776          & 0.815          \\
right pleural left area               & 0.248                & 0.614          & \textbf{0.946} & 0.734          & 0.819          \\
right retrocardiac area               & 0.701                & 0.917          & 0.954          & \textbf{0.968} & 0.962          \\
right rib area                        & 0.963                & \textbf{0.997} & \textbf{0.997} & \textbf{0.997} & \textbf{0.997} \\
right side                            & 0.928                & 0.978          & \textbf{0.979} & \textbf{0.979} & \textbf{0.979} \\
right upper area                      & \textbf{0.948}       & 0.906          & 0.940          & 0.928          & 0.934          \\
right upper lung area                 & 0.935                & 0.942          & 0.944          & 0.943          & \textbf{0.946} \\
scattered                             & 0.857                & \textbf{0.981} & 0.972          & \textbf{0.981} & 0.980          \\
scoliosis                             & 0.888                & 0.954          & 0.956          & 0.949          & \textbf{0.958} \\
severe                                & 0.886                & 0.973          & 0.971          & 0.969          & \textbf{0.974} \\
small                                 & 0.982                & \textbf{0.991} & \textbf{0.991} & 0.990          & \textbf{0.991} \\
small moderate                        & 0.811                & 0.967          & \textbf{0.968} & 0.965          & \textbf{0.968} \\
subtle                                & 0.955                & 0.996          & 0.994          & 0.996          & \textbf{0.997} \\
the apical area                       & \textbf{0.999}       & 0.726          & 0.924          & 0.876          & 0.884          \\
the left lower lung                   & 0.892                & 0.932          & 0.934          & 0.940          & \textbf{0.942} \\
the left lung base                    & 0.905                & \textbf{0.942} & \textbf{0.942} & 0.939          & \textbf{0.942} \\
the left middle lung                  & 0.885                & 0.930          & 0.931          & 0.932          & \textbf{0.936} \\
the left middle to lower lung         & \textbf{0.983}       & 0.855          & 0.711          & 0.839          & 0.810          \\
the left upper lung                   & 0.856                & 0.930          & \textbf{0.942} & 0.936          & \textbf{0.942} \\
the lower lung                        & 0.896                & 0.925          & 0.930          & 0.932          & \textbf{0.934} \\
the lower lungs                       & 0.911                & 0.929          & 0.926          & \textbf{0.936} & \textbf{0.936} \\
the lung bases                        & 0.906                & 0.935          & 0.937          & 0.936          & \textbf{0.938} \\
the middle lung                       & 0.877                & \textbf{0.889} & 0.822          & 0.884          & 0.881          \\
the right lower lung                  & 0.888                & 0.937          & \textbf{0.946} & 0.938          & 0.943          \\
the right lung base                   & 0.908                & 0.937          & \textbf{0.939} & 0.935          & 0.938          \\
the right middle lung                 & 0.912                & 0.933          & \textbf{0.936} & 0.933          & 0.934          \\
the right upper lung                  & 0.912                & 0.954          & \textbf{0.956} & 0.943          & \textbf{0.956} \\
the upper lung                        & 0.849                & 0.913          & 0.909          & 0.910          & \textbf{0.917} \\
the upper lungs                       & 0.937                & 0.930          & 0.980          & \textbf{0.988} & \textbf{0.988} \\
tortuosity of the descending aorta    & \textbf{0.897}       & 0.743          & 0.681          & 0.885          & 0.767          \\
tortuosity of the thoracic aorta      & 0.875                & 0.933          & 0.944          & 0.909          & \textbf{0.936} \\
upper area                            & 0.944                & 0.985          & 0.985          & 0.988          & \textbf{0.991} \\
upper lung area                       & 0.925                & 0.958          & 0.958          & 0.954          & \textbf{0.960} \\
vascular congestion                   & 0.933                & \textbf{0.944} & 0.941          & 0.937          & 0.941          \\
yes                                   & 0.949                & \textbf{0.991} & \textbf{0.991} & 0.990          & \textbf{0.991} \\
AUC-micro                             & 0.962                & \textbf{0.992} & \textbf{0.992} & \textbf{0.992} & \textbf{0.992} \\
AUC-macro                             & 0.848                & 0.901          & 0.905          & 0.909          & \textbf{0.912}
\end{longtable}

\end{document}